\documentclass[10pt,twocolumn,letterpaper]{article}

\usepackage[pagenumbers]{cvpr}

\usepackage{amsmath,amsfonts,bm}

\def\eqref#1{equation~\ref{#1}}

\def\1{\bm{1}}

\DeclareMathAlphabet{\mathsfit}{\encodingdefault}{\sfdefault}{m}{sl}
\SetMathAlphabet{\mathsfit}{bold}{\encodingdefault}{\sfdefault}{bx}{n}

\newcommand{\E}{\mathbb{E}}

\newcommand{\KL}{D_{\mathrm{KL}}}

\usepackage{capt-of}
\usepackage{tikz}
\usetikzlibrary{positioning,arrows.meta,fit,backgrounds,calc,shapes.geometric}
\usepackage{pgfplots}
\pgfplotsset{compat=1.18}
\usepackage{algorithm}
\usepackage{algpseudocode}
\usepackage{amsthm}
\usepackage{bm} 
\newtheorem{theorem}{Theorem}
\newtheorem{proposition}{Proposition}

\newtheorem{lemma}{Lemma}
\theoremstyle{definition}

\theoremstyle{remark}

\AtEndPreamble{\crefname{assumption}{Assumption}{Assumptions}\Crefname{assumption}{Assumption}{Assumptions}}

\newif\ifmarknew
\marknewfalse
\newcommand{\marknew}{\ifmarknew\color{red}\fi}

\usepackage[normalem]{ulem}
\newif\ifshowchanges
\showchangestrue

\definecolor{chpink}{RGB}{214,51,132}

\ifshowchanges\fi

\usepackage{longtable}
\usepackage{placeins}

\definecolor{cvprblue}{rgb}{0.21,0.49,0.74}
\usepackage[pagebackref,breaklinks,colorlinks,allcolors=cvprblue]{hyperref}

\def\paperID{*****}
\def\confName{CVPR}
\def\confYear{2027}

\title{{\marknew Generative Data Augmentation Without External Data:\\An Empirical and Theoretical Study on Deep Networks}}

\author{Daniel Kua, Yan Song\\
Department of Statistics, University of British Columbia\\
{\tt\small daniel.kua@stat.ubc.ca, yan.song@stat.ubc.ca}
}

\begin{document}

\newif\ifbloodsamples
\bloodsamplesfalse

\twocolumn[{%
\renewcommand\twocolumn[1][]{#1}%
\maketitle
\begin{center}
\centering
\includegraphics[width=\textwidth]{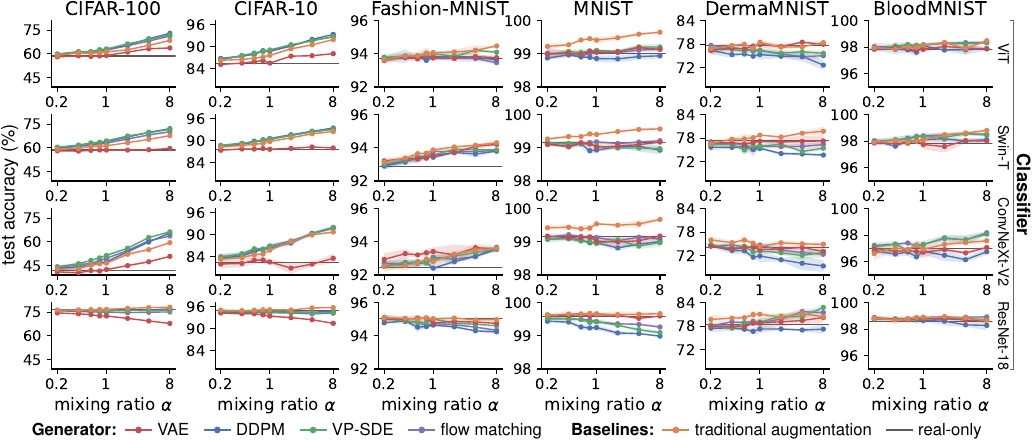}

\captionof{figure}{{\marknew \textbf{Accuracy moves in both directions as we add synthetic data.} Test accuracy (\%) against the mixing ratio $\alpha$, at real-data fraction $\rho = 1$. Rows: backbones. Columns: datasets, each with its own vertical scale. Lines: generators, traditional augmentation and the real-only classifier (black), as three-seed means with $\pm 1$ standard-deviation bands. Values: {Tabs.~A1 to~A3}.}}
\label{fig:interaction}
\end{center}%
}]

\begin{figure*}[!t]
\centering
\includegraphics[width=\textwidth]{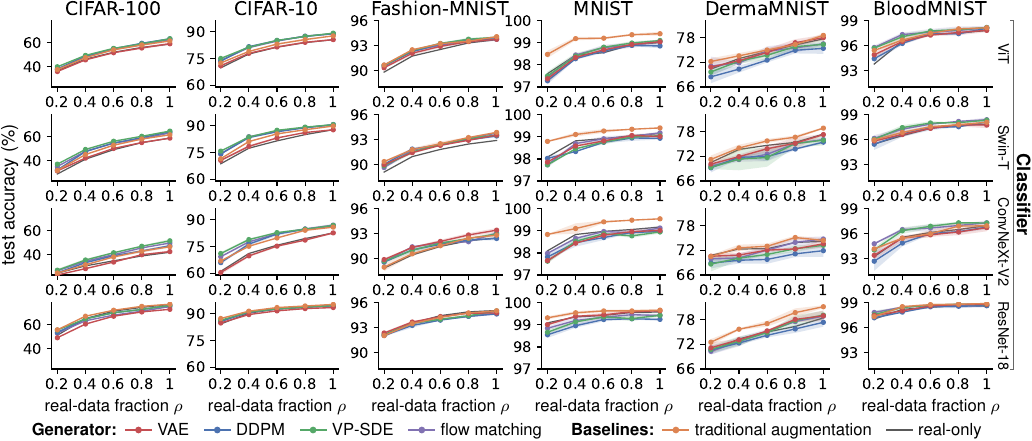}
\caption{{\marknew \textbf{Accuracy depends on how much real data the classifier has.} Test accuracy (\%) against the real-data fraction $\rho$ at $\alpha = 1.0$, with everything else as in \cref{fig:interaction}. Tables~A4 and~A5 give the values for the four benchmark datasets.}}
\label{fig:rho-blocks}
\end{figure*}

\begin{abstract}
{\marknew Generative data augmentation adds samples from generative models to a classifier's training set. Many influential studies use generators pre-trained on large external datasets, and so mix the value of synthetic data with knowledge transferred through pretraining. We study the case without external data, where each generator learns only from the classifier's own training set, and ask when and why adding generated samples to that training set raises or lowers the classifier's test accuracy. We train classifiers with four backbones on six datasets, two of them medical, adding samples from four generators. We bound the excess risk of a classifier trained with synthetic data, that is, how far its population risk can lie above the minimum risk within its hypothesis class. The bound charges synthetic data for the total variation distance between the real and generated distributions and credits synthetic data for a larger training set. The gain from adding synthetic data, the fall in population risk, is at most the excess risk of the real-only classifier, which trains on the real images alone. Read through the bound, the experiments suggest that synthetic data help most when our classifiers detect only a small discrepancy between the generated and real data and the real-only classifier has room to improve, and that synthetic data help little or hurt when the detected discrepancy is large. Generators with a larger lower bound on the total variation distance, a bound that we estimate from our own classifiers, tend to help less on five of the six datasets, and on the CIFAR datasets the most accurate backbone gains the least. The bound gives these directions but not the size of the largest gains, most of which traditional augmentation also reaches at the same number of optimizer steps.}

\end{abstract}

\newpage

\section{Introduction}

\label{sec:intro}

{\marknew Synthetic data augmentation expands a training set with artificially created samples, used alongside or in place of real data, to address data scarcity \citep{mumuni2024survey}. Traditional data augmentation modifies existing real images, for instance by flipping, cropping or rotating them, to improve generalization and reduce overfitting \citep{mumuni2024survey}. Generative data augmentation (GDA), the branch of synthetic data augmentation we study, creates new images with generative models, such as generative adversarial networks (GANs), variational autoencoders (VAEs) and diffusion models, which learn the distribution of real data \citep{mumuni2024survey, wang2025closedset}. Many influential studies train image classifiers on images from text-to-image models,
such as Imagen, GLIDE and Stable Diffusion, pre-trained on large external datasets \citep{azizi2023synthetic, he2023synthetic, zhou2023using, fan2024scaling}. These studies mix the value of the synthetic data with knowledge transferred through pretraining, as real images retrieved from the same pretraining data can match or beat the synthetic ones on natural-image benchmarks \citep{burg2023image, geng2024unmet}.}

\begin{figure*}[!t]
\centering
\includegraphics[trim=2pt 2pt 2pt 2pt,clip]{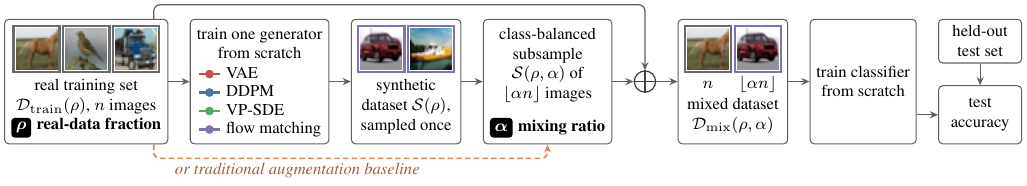}%
\caption{{\marknew \textbf{The generator never sees data beyond the classifier's training set.} Black tags mark the two factors we vary, and $\oplus$ is the union of \cref{eq:mix}. The last box gives top-1 test accuracy.}}
\label{fig:pipeline}
\end{figure*}

{\marknew GDA without external data is particularly relevant when privacy or licensing restrictions prevent access to other datasets and when collecting and annotating new real images is difficult and expensive, as in medical imaging \citep{groger2025medical}. However, the effects of synthetic data on downstream classification remain unclear. Previous studies have reported both gains \mbox{\citep{wang2025closedset}} and losses \mbox{\citep{shmelkov2018good}} in classification accuracy. \citet[Section~4.2]{zheng2023toward} observe both with GANs on CIFAR-10: synthetic images can improve accuracy without standard augmentation but may reduce it with standard augmentation. Similarly, our experiments show that test accuracy can increase or decrease as the amount of synthetic data (\cref{fig:interaction}) and the size of the real training set (\cref{fig:rho-blocks}) vary, depending on the classifiers and generative models used. These mixed results motivate further empirical and theoretical investigation of how synthetic images affect classifiers when generators learn only from the classifiers' own training data (\cref{fig:pipeline}).}

{\marknew In this paper, we study GDA without external data for image classification, through both experiments and theory. Each generator learns only from the classifier's real training set, and each classifier trains from scratch on that set together with the generator's samples. We compare four deep generative models (DGMs), namely variational autoencoders, denoising diffusion probabilistic models, score-based models and flow matching, and four classifier architectures, namely ResNet-18, a Vision Transformer (ViT), Swin-T and ConvNeXt-V2, on six image datasets, two of them medical image datasets with imbalanced classes (\cref{sec:framework}). We vary the mixing ratio $\alpha$, the number of synthetic images added per real image, and the real-data fraction $\rho$, the share of the real training set we keep. To explain when and why synthetic data raise or lower test accuracy, we extend the stability bound of \citet[Theorem~3.1]{zheng2023toward} from the generalization error to the excess risk, bounding how far the population risk of a classifier trained with synthetic data can lie above the minimum risk within its hypothesis class (\cref{sec:tv}). The bound charges synthetic data for the total variation distance between the real and generated distributions and credits them for a larger training set. A mixture's gain over the real-only classifier is at most that classifier's excess risk} (\cref{eq:cap}). We read the accuracy curves through these terms and estimate a lower bound on this distance from our own classifiers (\cref{sec:experiments}). We summarize our main contributions as follows:

\begin{itemize}
    \item {\textbf{A comprehensive empirical study of GDA without external data.} We systematically evaluate GDA using six datasets, four deep generative models (VAE, DDPM, VP-SDE, and flow matching), and four classifiers (ResNet-18, ViT, Swin-T, and ConvNeXt-V2), without external pretraining data. By varying the amount of real training data and the proportion of synthetic data, we examine their effects on classification performance and compare against traditional augmentation.}

\item {\textbf{A theoretical analysis of population excess risk.} Building on \citet{zheng2023toward}, we derive an upper bound on the population excess risk under GDA. The analysis identifies key factors, including distribution discrepancy, mixing ratio, sample size, classifier stability, and dependence between real and generated data. We further relate the bound to the potential accuracy gain or loss relative to real-only training.}

\item {\textbf{Empirical findings and theoretical insights.} Our results suggest that generators with a larger estimated lower bound on the distribution discrepancy tend to yield smaller accuracy gains, while more accurate real-only classifiers generally benefit less, particularly on the CIFAR datasets. These patterns are consistent with our theoretical analysis, although the bound does not guarantee accuracy improvements. Traditional augmentation also remains competitive, often achieving comparable or better performance without training a generative model.}
\end{itemize}

\section{Related work}
\label{sec:background}

\subsection{\texorpdfstring{Generative data augmentation}{Generative data augmentation}}
\label{sec:rw-gda}

GDA with external data draws its images from generators pretrained on large external datasets. Images from text-to-image models such as Imagen, GLIDE and Stable Diffusion can raise accuracy on ImageNet and other benchmarks \citep{azizi2023synthetic, he2023synthetic, sariyildiz2023fake, fan2024scaling}, with gains that depend on the backbone \citep{azizi2023synthetic, he2023synthetic, oehri2024genformer}. These gains mix the value of the synthetic images with knowledge from the pretraining data, since real images retrieved from that data match or beat them \mbox{\citep{burg2023image, geng2024unmet}}.

Without external data, the generator learns only from the classifier's training set, and the results are mixed. Samples from generative adversarial networks (GANs) lower accuracy on CIFAR-10 and CIFAR-100 \citep{shmelkov2018good} and on ImageNet \citep{ravuri2019cas}, yet help a liver-lesion classifier \citep{fridadar2018gan}, and their value for medical image segmentation depends on the amount of real data \citep{bowles2018gan}. Diffusion models trained from scratch help on CIFAR-10, BloodMNIST and ImageNet-100, but less than the same number of real images \citep{wang2025closedset}, and replacing many real images with generated ones at a fixed training-set size lowers accuracy \citep{lesort2019training}. \citet{wangr2024improving} trace the weak value of generated CIFAR-10 images to rare modes that the generators omit, with classifiers pretrained on ImageNet, and \citet{an2025generative} raise accuracy on CIFAR-10 and ImageNet-1K by aligning and reweighting in-domain diffusion samples.

Theory explains such results through generalization bounds. Algorithmic stability bounds the generalization error of a learning algorithm by how much its output changes when we replace one training point \citep{bousquet2002stability, bousquet2020sharper}, and stochastic gradient descent run for few steps is stable \citep{hardt2016train}. \citet{zheng2023toward} extend stability bounds to real data mixed with samples from a generator fitted to the real data, adding a total variation term and a term for the stability of the generator. On CIFAR-10, the authors estimate the generalization error by the gap between training and test cross-entropy \citep[Appendix~D.7]{zheng2023toward}. The cost of training on another distribution is classical in domain adaptation \citep{bendavid2010theory}, and related bounds use total variation for generated data in contrastive learning \citep{wang2024generated} and Wasserstein distances in kernel ridge regression \citep{shidani2026beyond} and for tabular data \citep{abeykoon2026reliability}. Such bounds are often vacuous on deep networks \citep{dziugaite2017computing, nagarajan2019uniform}, and many bound-based measures correlate poorly with generalization \citep{jiang2020fantastic}, so we use our excess-risk extension to explain trends, not to give a numerical guarantee.

Like \citet{shmelkov2018good} and \citet{wang2025closedset}, we retrain each generator on every real-data fraction. Unlike them, we compare four generators, convolutional and transformer classifiers, and natural and medical datasets under one protocol, and we explain the accuracy trends with an excess-risk bound and a lower bound on the total variation distance between the real and generated distributions.

\subsection{\texorpdfstring{Deep generative models}{Deep generative models}}
\label{sec:rw-dgm}

We use four deep generative models, which differ in how they learn the data distribution and how they sample. Variational autoencoders (VAEs) \citep{kingma2014auto, rezende2014stochastic} train an encoder and a decoder by maximizing a lower bound on the likelihood and sample in one decoder pass, but produce blurrier samples than the other three generators (\cref{fig:families}). Denoising diffusion probabilistic models (DDPMs) \citep{ho2020denoising, nichol2021improved, ho2022classifier} learn to reverse a fixed process that adds Gaussian noise step by step, and reach high fidelity \citep{dhariwal2021diffusion} at the cost of many network evaluations per sample, which faster samplers reduce \citep{song2021denoising}. Score-based models \citep{song2019generative, song2021score} learn the gradient of the log density of noised data and sample by solving a reverse-time stochastic differential equation. We use the variance-preserving one (VP-SDE), whose discretization gives the DDPM \citep{song2021score, karras2022elucidating}. Flow matching \citep{lipman2023flow, albergo2023building, liu2023flow} instead regresses a velocity field that carries noise to data along simple paths and samples by solving an ordinary differential equation, often in fewer steps \citep{liu2023flow}. We train each generator to condition on the class label, so every synthetic image has a label (\cref{sec:instantiations}).

\subsection{\texorpdfstring{Classification models}{Classification models}}
\label{sec:rw-cls}

We train four classifier architectures with different inductive biases, two convolutional networks and two vision transformers. ResNet-18 \citep{he2016deep} stacks convolutional blocks with residual connections. ConvNeXt-V2 \citep{woo2023convnext} modernizes the convolutional network with large depthwise kernels and layer normalization and adds global response normalization. ViT \citep{dosovitskiy2021vit} splits an image into patches and lets every patch attend to every other, and with few built-in assumptions about images it needs more training data than a convolutional network. Swin-T \citep{liu2021swin} restricts attention to shifted local windows and builds a hierarchy of feature maps, which restores some locality. These differences matter here, because the gain from synthetic images depends on the backbone \citep{azizi2023synthetic, he2023synthetic, oehri2024genformer}.

\section{Method}
\label{sec:framework}

\leavevmode{\marknew This section describes the experimental framework, including the datasets, generators, and classifiers. We train neither the generators nor the classifiers on external data. Each comparison varies one factor and fixes the others, up to the couplings in \cref{app:caveats}.

\subsection{\texorpdfstring{Datasets and experimental protocol}{Datasets and experimental protocol}}
\label{sec:protocol}

\begin{figure*}[!t]
\centering
\includegraphics[trim=2pt 2pt 2pt 2pt,clip]{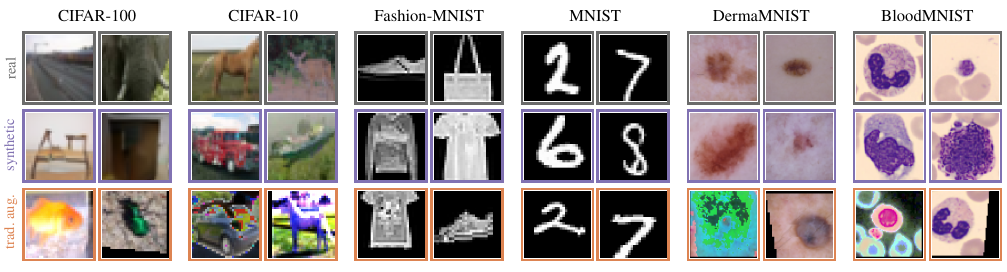}%
\caption{{\marknew \textbf{Real, synthetic and traditionally augmented images of the six datasets.} Synthetic: samples from the flow-matching generator. Trad.\ aug.: images} from the traditional augmentation pool that the classifiers train on.}
\label{fig:datasets}
\end{figure*}

\paragraph{Datasets.}
\leavevmode{\marknew We use six datasets, four benchmarks with varying levels of classification difficulty and two medical image datasets with imbalanced classes. The benchmarks are CIFAR-100, CIFAR-10 \citep{krizhevsky2009learning}, Fashion-MNIST \citep{xiao2017fashion} and MNIST \citep{lecun1998gradient} (\cref{fig:datasets}, with more samples in \cref{app:float-samples}). Fashion-MNIST shares MNIST's format of $28 \times 28$ grayscale images, ten classes and 60{,}000 training images, but presents a more challenging classification task. The medical datasets, from MedMNIST+ \citep{yang2023medmnist}, are both $64 \times 64$ RGB. After the validation split,} DermaMNIST has seven imbalanced classes, 6{,}306 training images and 72 images in the rarest class. BloodMNIST has eight classes, 10{,}763 training images and a 2.7 to 1 imbalance.

\paragraph{Training mixtures.}

\leavevmode{\marknew Each training mixture has two parameters: the real-data fraction $\rho$, controlling the proportion of real training data retained for training the generator and classifier, and the mixing ratio $\alpha$, controlling the number of synthetic images added relative to the retained real images (\cref{sec:intro}).

The full real training set $\mathcal{D}_{\mathrm{train}}(1)$ is the official training set minus the validation split, and it holds $N$ images. The real training set $\mathcal{D}_{\mathrm{train}}(\rho)$ keeps the fraction $\rho$ of each class and holds $n \approx \rho N$ images. We train each generator on $\mathcal{D}_{\mathrm{train}}(\rho)$ and sample a  class-balanced synthetic dataset $\mathcal{S}(\rho)$. For given $\alpha$ and $\rho$, a class-balanced subsample $\mathcal{S}(\rho, \alpha)$ of $\mathcal{S}(\rho)$ joins the real training set. The training mixture is then}
\begin{equation}
\label{eq:mix}
\begin{gathered}
\mathcal{D}_{\mathrm{mix}}(\rho, \alpha) \;=\; \mathcal{D}_{\mathrm{train}}(\rho) \,\cup\, \mathcal{S}(\rho, \alpha),\\
|\mathcal{S}(\rho, \alpha)| \;=\; \big\lfloor \alpha n \big\rfloor .
\end{gathered}
\end{equation}
{\marknew The grid spans $\alpha \in \{0.2, 0.4, 0.6, 0.8, 1.0, 2.0, 4.0, 8.0\}$ and $\rho \in \{0.2, 0.4, 0.6, 0.8, 1.0\}$, and the design is not fully factorial (Tab.~A8). Unless a table or figure says otherwise, every experiment uses $\rho = 1$. Every run uses the base augmentation of Tab.~A8. We report means over three run seeds, which vary classifier training and synthetic data sub-sampling, while we train each generator once per dataset and real-data fraction} (\cref{app:impl}).

\paragraph{Design constraints.}
\leavevmode{\marknew Each generator sees only $\mathcal{D}_{\mathrm{train}}(\rho)$, and we retrain it at every $\rho$. Every generator on a dataset receives the same number of image updates. \Cref{app:impl,app:hparams} give the training details and hyperparameters. The classifier means a backbone with its fixed training recipe. We fix its budget in epochs, so raising $\alpha$ also raises the number of optimizer steps.}

\subsection{\texorpdfstring{Generative models and traditional augmentation}{Generative models and traditional augmentation}}
\label{sec:instantiations}

{\marknew We train the four generators of \cref{sec:rw-dgm} from scratch, each with its standard objective (\cref{app:generators}). DDPM, VP-SDE and flow matching share one class-conditional U-Net, to control for differences in network capacity, while the VAE has its own convolutional encoder and decoder. All four train with AdamW at a constant learning rate of $2 \times 10^{-4}$ and a batch size of 128, and sample with an exponential moving average of their weights. Each trains for 9.0 million image updates at $\rho = 1$, or 1.6 million on MNIST ({Tab.~A12}). To draw $\mathcal{S}(\rho)$, DDPM runs 250 DDIM steps, the VP-SDE 500 predictor-corrector steps and flow matching a 50-step midpoint solver, while the VAE decodes in one pass ({Tab.~A13})}.  

\paragraph{Sample quality.}
 We measure the sample quality of each synthetic dataset by its Fr\'echet Inception Distance (FID) \citep{heusel2017gans} to the generator's own training set. FID compares the two sets in the feature space of an InceptionV3 network (\cref{app:impl}) and measures the sample quality each DGM reaches under our fixed training budget and gives a common metric for how closely DGM samples resemble the real training images.  The bound of \cref{sec:tv} measures the distance between the real and generated distributions by $d_{\mathrm{TV}}(P, G)$, which we cannot compute directly on images and bound only from below (\cref{sec:results-read}).  Although FID does not estimate $d_{\mathrm{TV}}(P, G)$,  we use FID to interpret the classification results in \cref{sec:results-read}. By FID, we call DDPM, VP-SDE and flow matching higher-fidelity generators and the VAE a lower-fidelity generator under our settings (\cref{fig:families}). The difference is substantial on every dataset except DermaMNIST. On CIFAR-10, the VAE has an FID of 113.9, compared with at most 11.0 for the other three generators. Across the four benchmark datasets, the VAE's FID is at least 2.5 times the highest FID among the other three (\cref{tab:pools}).

\paragraph{The traditional augmentation baseline.}
\leavevmode{\marknew \Cref{fig:interaction,fig:rho-blocks} also show a traditional augmentation baseline, which adds transformed real images in place of generated ones. For each class, we draw images from $\mathcal{D}_{\mathrm{train}}(\rho)$ uniformly with replacement and transform each draw once {(\cref{fig:datasets})}. The transform is RandAugment \citep{cubuk2020randaugment} with two operations at magnitude 9 on the RGB datasets, and a random rotation of up to 10 degrees with a random shift of up to a tenth of the image size on the $28 \times 28$ datasets. A random crop after 4-pixel padding and a random horizontal flip follow on every dataset except MNIST (\cref{app:generators}). The comparison with the generators is fair, because the baseline differs from a generated mixture only in where the added images come from. The baseline uses the same real training set $\mathcal{D}_{\mathrm{train}}(\rho)$ as the generators and no other data and replaces $\mathcal{S}(\rho)$ in \cref{eq:mix}, so a mixture adds the same number $\lfloor \alpha n \rfloor$ of class-balanced images, and the classifier trains on it with the same recipe, base augmentation, checkpoint selection and run seeds. The baseline fixes its transformed images before training, like a synthetic dataset, while base augmentation transforms each training image anew whenever the classifier loads it.

\subsection{\texorpdfstring{Classifier training and evaluation}{Classifier training and evaluation}}
\label{sec:classifiers}

{\marknew Apart from the InceptionV3 behind FID, we use no external pretrained weights and train the four backbones of \cref{sec:rw-cls} from scratch in every configuration, with a compact ViT and ConvNeXt-V2 in its Femto size (Tab.~A11). Each backbone has one fixed recipe, identical for every generator and mixing ratio. ResNet-18 trains with SGD with momentum and the other three with AdamW, all with a cosine learning-rate schedule, a batch size of 128 and an unweighted cross-entropy loss. Training lasts 100 epochs, except 200 for the ViT and 30 for ResNet-18 and Swin-T on MNIST. After each epoch, we evaluate the classifier on the validation set, and we keep the checkpoint with the best validation accuracy. Our evaluation metric is the top-1 accuracy of the kept checkpoint on the official test set, which never informs a choice.
The theory of \cref{sec:tv}, which bounds the population excess risk of a classifier trained on the mixture, helps interpret the results.

\begin{figure}[!t]
\centering
\includegraphics[trim=2pt 2pt 2pt 2pt,clip]{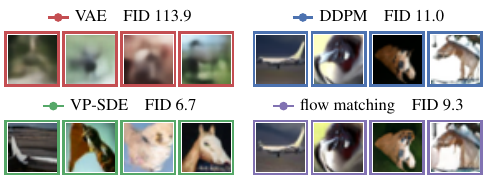}%
\caption{\textbf{At the same training budget, the VAE is the lower-fidelity generator.} CIFAR-10 samples from the same sheet positions for every generator.}
\label{fig:families}
\end{figure}

\section{\texorpdfstring{Excess risk with synthetic training data}{Excess risk with synthetic training data}}
\label{sec:tv}

To investigate how GDA affects classification performance, we extend \citet[Theorem~3.1]{zheng2023toward} to bound the population excess risk of a classifier trained on the mixture. This excess risk measures how much the classifier's population risk exceeds the minimum risk within its hypothesis class. Table A15 lists the notation.

\paragraph{Setting.} \leavevmode{\marknew  The full real training set $\mathcal{D}_{\mathrm{train}}(1)$ holds $N$ labeled images, which we model as independent draws from the distribution $P$ of labeled images. The training set $\mathcal{D}_{\mathrm{train}}(\rho)$ contains $n = \rho N$ of these images, up to rounding, and we use it to train both the generator and the classifier. A generator fitted to $\mathcal{D}_{\mathrm{train}}(\rho)$ yields a distribution $G_{\rho}$, written $G$ when $\rho$ is fixed. We draw about $\alpha n$ samples from $G_\rho$ and add them to the real training set $\mathcal{D}_{\mathrm{train}}(\rho)$. The corresponding mixture distribution is
$P_{\alpha} = \frac{1}{1+\alpha} P + \frac{\alpha}{1+\alpha} G_{\rho}$. 

\leavevmode{\marknew For a loss $\ell$ bounded in $[0, M]$, we define the population risk of a classifier $f$ under a distribution $Q$ as $R_{Q}(f) = \E_{(x, y) \sim Q}\, \ell(f(x), y)$. The empirical risk on the training mixture is $\widehat{R}_{P_{\alpha}}(f)=\frac{1}{|\mathcal{D}_{\mathrm{mix}}(\rho,\alpha)|}\sum_{(x_i,y_i)\in\mathcal{D}_{\mathrm{mix}}(\rho,\alpha)}\ell(f(x_i),y_i)$}. Under the 0-1 loss, $M = 1$, and $R_{P}(f)$ is the population misclassification risk of $f$, which we estimate using the test-set error $\widehat{R}^{\mathrm{test}}_{P}(f)$ (\cref{app:tv-sim}). Let $f^{\star}_{P}=\arg\min_{f\in\mathcal{F}} R_P(f)$ denote the population-risk minimizer. The training algorithm $\mathcal{A}$ returns $\hat{f}_{\alpha}$ when trained on the mixture and $\hat{f}_{0}$ when trained on real data only. Both classifiers depend on $\rho$, which we leave out of the notation. Since we train on the cross-entropy and keep the best-validation checkpoint, $\hat{f}_{\alpha}$ need not minimize the empirical 0-1 risk 
$\widehat{R}_{P_{\alpha}}$.}

\paragraph{Key factors in the population excess-risk bound.} \leavevmode{\marknew The theoretical bound involves three key quantities. First, algorithm $\mathcal{A}$ is called $\beta_{m}$-uniformly stable if replacing one of its $m$ training points changes the loss of its output on any labeled image by at most $\beta_{m}$ \citep[Definition~2.1]{zheng2023toward}. Second, we measure the discrepancy between the real and generated distributions by the total variation distance $d_{\mathrm{TV}}(P, G_{\rho}) = \sup_{A} | P(A) - G_{\rho}(A) |$, where the supremum runs over all measurable events $A$. For any classifier $f$ and any loss bounded in $[0,M]$, this distance bounds the difference between the population risks under $P$ and $G_\rho$ (\cref{app:tv-setting}),
\begin{equation}
\label{eq:tv-gap}
| R_{P}(f) - R_{G_{\rho}}(f) | \le M\, d_{\mathrm{TV}}(P, G_{\rho}) .
\end{equation}
Third, since we fit the generator to $\mathcal{D}_{\mathrm{train}}(\rho)$, $G_{\rho}$ depends on the real training data. The quantity $\mathcal{T}$ measures the sensitivity of the generated sample distribution to changes in the real training data. Specifically, it is the largest change in total variation between the distributions of the} $\alpha \rho N$ generated samples when we replace one real training observation. \citep[Theorem~3.1]{zheng2023toward}.

\begin{figure*}[t]
\centering
\includegraphics[width=\textwidth]{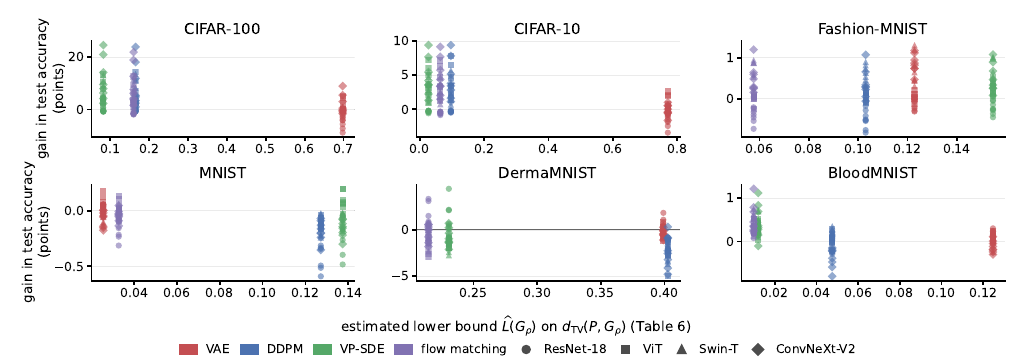}
\caption{{\marknew \textbf{Generators with a larger estimated lower bound help less on five of the six datasets.} Gain in test accuracy $\widehat{\mathrm{Err}}(0) - \widehat{\mathrm{Err}}(\alpha)$ of each mixture at $\rho = 1$, in points, against $\widehat{L}(G_{\rho})$, the largest gap between a classifier's test error and its error on images from $G_{\rho}$.}}
\label{fig:cost}
\end{figure*}

\paragraph{Excess-risk bound.} \leavevmode{\marknew Following \citet[Section~2.2]{zheng2023toward}, we first assume that, conditional on $\mathcal{D}_{\mathrm{train}}(\rho)$ and the generator's training randomness, the generated observations are independent draws from $G_{\rho}$. Our experimental protocol instead uses fixed per-class counts, under which the bound still holds (\cref{app:caveats}). If $\mathcal{A}$ is $\beta_{(1+\alpha) \rho N}$-uniformly stable with respect to the bounded loss $\ell$, then for any $\alpha \ge 0$ and $\delta \in (0, 1/2]$, with probability at least $1 - \delta$,
\begin{equation}
\label{eq:tv-stab-dep}
\begin{aligned}
&R_{P}(\hat{f}_{\alpha}) - R_{P}(f^{\star}_{P}) \lesssim \frac{2\alpha}{1+\alpha}\, M\, d_{\mathrm{TV}}(P, G_{\rho}) \\
&\quad + M \sqrt{\frac{\log(1/\delta)}{(1+\alpha) \rho N}} + \frac{\sqrt{\alpha \rho N}}{1+\alpha}\, \beta_{(1+\alpha) \rho N} \sqrt{\log\tfrac{1}{\delta}} \\
&\quad + \frac{\log(\rho N) + \alpha \log(\alpha \rho N)}{1+\alpha}\, \beta_{(1+\alpha) \rho N} \log\tfrac{1}{\delta} \\
&\quad + \frac{M\, \mathcal{T} \log(\rho N)}{1+\alpha} \log\tfrac{1}{\delta} + \widehat{R}_{P_{\alpha}}(\hat{f}_{\alpha}) ,
\end{aligned}
\end{equation}
where we take $\alpha \log(\alpha \rho N) = 0$ at $\alpha = 0$. The bound consists of five components:}\par
\begin{itemize}
\item {\marknew \textbf{Distribution-discrepancy term}, $\frac{2\alpha}{1+\alpha} M\, d_{\mathrm{TV}}(P, G_{\rho})$, the potential cost of training on generated observations from $G_\rho$ rather than real observations from $P$. This term increases with $\alpha$ at fixed $G_\rho$ and with the discrepancy between the two distributions.}

\item {\marknew \textbf{\textbf{Sample-size} term}, $M \sqrt{\log(1/\delta) / ((1+\alpha) \rho N)}$, which decreases as $\alpha$ grows at fixed $\rho$ and $N$, reflecting a potential statistical benefit of additional samples.}
\item {\marknew \textbf{Stability terms}, the two terms involving $\beta_{(1+\alpha) \rho N}$, the stability of the training algorithm $\mathcal{A}$, which depends on how we train the classifier \citep{bousquet2002stability, hardt2016train}. Under the 0-1 loss, replacing a single training point can change a prediction, and the trivial stability bound $\beta_m \leq 1$ yields large stability terms, so we do not interpret them quantitatively.}

\item {\marknew \textbf{Dependence term}, the term involving $\mathcal{T}$, which characterizes the sensitivity of the learned generator distribution to changes in its training data \citep[Section~3.1]{zheng2023toward}.} This term vanishes when the generator is independent of the real training data.
\item {\marknew \textbf{Empirical-risk term}, $\widehat{R}_{P_{\alpha}}(\hat{f}_{\alpha})$, which accounts for the empirical loss of the trained classifier on the mixture. Under the 0-1 loss, it is zero if the classifier correctly classifies every training observation.}
\end{itemize}

\leavevmode{\marknew \Cref{app:tv-dep-case} proves the bound by applying \citet[Theorem~3.1]{zheng2023toward} twice.}

\paragraph{How $\alpha$, $\rho$ and the distribution shift affect the bound.} \leavevmode{\marknew The right side of \cref{eq:tv-stab-dep} grows with the distribution shift $d_{\mathrm{TV}}(P, G_{\rho})$ through the distribution-discrepancy term, so a generator farther from $P$ raises the bound. Raising $\alpha$ increases the distribution-discrepancy term, since its weight $\frac{2\alpha}{1+\alpha}$ rises from $0$ toward $2$. Raising $\alpha$ also lowers the sample-size term, since the training set grows to $(1+\alpha) \rho N$ points, and the fall is larger when $\rho N$ is small. A smaller $\rho$ therefore enlarges the fall of the sample-size term, but we then fit $G_{\rho}$ to fewer real images, which can raise $d_{\mathrm{TV}}(P, G_{\rho})$. If we temporarily ignore the stability and empirical-risk terms, the remaining upper bound decreases relative to $\alpha=0$ when the reduction in the sample-size term exceeds the distribution-discrepancy and dependence terms. The reduction is at most of order $1/\sqrt{\rho N}$.  Thus, for fixed $\alpha > 0$, $d_{\mathrm{TV}}(P,G_\rho)$ and $\mathcal T\log(\rho N)$ must be small relative to $1/\sqrt{\rho N}$ for this simplified bound to decrease (compare \citet[Corollary~3.1]{zheng2023toward}). This comparison concerns the theoretical upper bound and does not guarantee an improvement in classification accuracy.}

\paragraph{From the bound to what we measure.} \leavevmode{\marknew We cannot evaluate the left side of \cref{eq:tv-stab-dep} directly, since the minimum population risk $R_{P}(f^{\star}_{P})$ in $\mathcal{F}$ is unknown. We therefore compare the mixture with the real-only classifier through the change $\mathrm{Err}(\alpha) - \mathrm{Err}(0)$, where $\mathrm{Err}(\alpha) = R_{P}(\hat{f}_{\alpha})$ is the population risk of the classifier trained at $\alpha$. Like the left side of \cref{eq:tv-stab-dep}, the change puts the mixture first. Since the real-only classifier $\hat{f}_{0}$ lies in $\mathcal{F}$, its population risk is at least the minimum, $\mathrm{Err}(0) = R_{P}(\hat{f}_{0}) \ge R_{P}(f^{\star}_{P})$. Subtracting $\mathrm{Err}(0)$ instead of $R_{P}(f^{\star}_{P})$ can therefore only lower the left side of \cref{eq:tv-stab-dep}, so the change is a lower bound on that left side,
\begin{equation}
\label{eq:cap}
\begin{aligned}
\mathrm{Err}(\alpha) - \mathrm{Err}(0) &= R_{P}(\hat{f}_{\alpha}) - R_{P}(\hat{f}_{0}) \\
&\le R_{P}(\hat{f}_{\alpha}) - R_{P}(f^{\star}_{P}) .
\end{aligned}
\end{equation}
To estimate the change $\mathrm{Err}(\alpha) - \mathrm{Err}(0)$, we use $\widehat{\mathrm{Err}}(\alpha) - \widehat{\mathrm{Err}}(0)$, where $\widehat{\mathrm{Err}}(\alpha)$ is the test error $\widehat{R}^{\mathrm{test}}_{P}(\hat{f}_{\alpha})$ on the held-out test set, averaged over the three run seeds of \cref{sec:protocol}. By \cref{eq:cap}, \cref{eq:tv-stab-dep} also bounds the change from above, so a larger distribution-discrepancy term allows a larger rise in risk. The gain $\mathrm{Err}(0) - \mathrm{Err}(\alpha)$ is the negative of the change, and since $\mathrm{Err}(\alpha) \ge R_{P}(f^{\star}_{P})$ as well, the gain is at most the excess risk of the real-only classifier. A classifier close to the best in its class therefore has little to gain. \Cref{sec:experiments} uses these results to interpret the observed changes in classification performance, while recognizing that comparing upper bounds does not determine which classifier achieves lower population risk.

\section{Experiments}
\label{sec:experiments}

\leavevmode{\marknew We examine two relationships motivated by the theoretical analysis in \cref{sec:tv} with the classifiers of \cref{fig:interaction}. \Cref{sec:results-read} shows how we estimate classification gains and use our classifiers to estimate a lower-bound quantity for the distribution shift. \Cref{sec:results-generator} examines whether generators with a larger estimated lower bound on the shift help less, and \cref{sec:results-backbone} whether more accurate real-only classifiers, with less room to improve, gain less. \Cref{fig:cost} shows both effects together.}

\subsection{\texorpdfstring{{Measuring accuracy gain and distribution discrepancy}}{measuring accuracy gain and distribution discrepancy}}
\label{sec:results-read}

\paragraph{A gain is a fall in excess risk.} \leavevmode{\marknew Within one panel of \cref{fig:interaction}, $P$ and $\mathcal{F}$ are fixed, so the change $\mathrm{Err}(\alpha) - \mathrm{Err}(0)$ is the excess risk of $\hat{f}_{\alpha}$ minus that of $\hat{f}_{0}$. We estimate the gain $\mathrm{Err}(0) - \mathrm{Err}(\alpha)$ by $\widehat{\mathrm{Err}}(0) - \widehat{\mathrm{Err}}(\alpha)$, which is also the gain in test accuracy, since test accuracy is one minus the test error. \Cref{fig:cost} plots this estimated gain in points, and a point above zero means that the mixture has a higher test accuracy than the real-only classifier. In \cref{fig:interaction}, the vertical distance of a mixture's curve above the black real-only line represents the same estimated gain.}

\paragraph{Our classifiers bound the distribution shift from below.} \leavevmode{\marknew By \cref{eq:tv-gap} with $M = 1$, every classifier $f$ satisfies $| R_{P}(f) - R_{G_{\rho}}(f) | \le d_{\mathrm{TV}}(P, G_{\rho})$. The largest of these gaps over our classifiers of \cref{fig:interaction} that train on real data alone or with that generator, $L(G_{\rho}) = \max_{f} | R_{P}(f) - R_{G_{\rho}}(f) |$, is therefore a lower bound on $d_{\mathrm{TV}}(P, G_{\rho})$.  Since the population risks are unknown, we estimate this lower-bound quantity using the empirical risks, $\widehat{L}(G_{\rho}) = \max_{f} | \widehat{R}^{\mathrm{test}}_{P}(f) - \widehat{R}^{\mathrm{test}}_{G_{\rho}}(f) |$, where $\widehat{R}^{\mathrm{test}}_{G_{\rho}}(f)$, the error of $f$ on about 5{,}000 fresh draws from $G_{\rho}$, estimates $R_{G_{\rho}}(f)$ ({Tab.~A6}).}

\subsection{\texorpdfstring{Estimated distribution discrepancy and classification gain}{Estimated distribution discrepancy and classification gain}}
\label{sec:results-generator}

\leavevmode{\marknew Since $d_{\mathrm{TV}}(P, G_{\rho}) \ge L(G_{\rho})$, the distribution-discrepancy term of \cref{eq:tv-stab-dep} is at least $\frac{2\alpha}{1+\alpha} M\, L(G_{\rho})$. A large $\widehat{L}(G_{\rho})$ therefore forces a large distribution-discrepancy term in the upper bound, although this does not necessarily imply a smaller classification gain. We investigate whether the estimated discrepancy $\widehat L(G_\rho)$ is associated with classification gain. In \cref{fig:cost}, each generator forms one column of points at its $\widehat{L}(G_{\rho})$, and columns further right tend to sit lower on five of the six datasets. On CIFAR-100, DDPM, the VP-SDE and flow matching sit at the left and reach far above zero, while the VAE sits far to the right and spreads around zero. In \cref{fig:interaction}, the curves of these three generators rise above the black real-only line for ViT, Swin-T and ConvNeXt-V2, while the VAE curves stay much closer to that line. CIFAR-10 shows the same order with smaller gains, and on both CIFAR datasets the points for these three generators lie above the corresponding VAE points with the same backbone and mixing ratio. On Fashion-MNIST the four columns lie close together and their points stay near zero, so $\widehat{L}(G_{\rho})$ cannot clearly distinguish the generators, and their curves stay near the real-only line. On MNIST, DDPM and the VP-SDE sit to the right of the VAE and flow matching and also lowest, mostly below zero, and in \cref{fig:interaction} their curves fall slightly below the real-only line. On DermaMNIST, DDPM sits furthest right and lowest, mostly below zero, but the VAE at almost the same position stays near zero, no lower than the VP-SDE and flow matching on the left. On BloodMNIST, the VP-SDE and flow matching sit at the left and almost all above zero, while DDPM and the VAE, further right, sit near zero.}

\subsection{\texorpdfstring{More accurate backbones tend to gain less, particularly on CIFAR}{More accurate backbones tend to gain less, particularly on CIFAR}}
\label{sec:results-backbone}

\leavevmode{\marknew The gain is at most the excess risk of the real-only classifier (\cref{sec:tv}) and hence at most its population risk, so a more accurate backbone has a smaller upper limit on its possible gain. In \cref{fig:cost} the marker shape indicates the backbone, and in \cref{fig:interaction} the black line of each row indicates that backbone's real-only accuracy. On CIFAR-100, ResNet-18 is the most accurate real-only classifier and ConvNeXt-V2 the least, and within each column corresponding to DDPM, the VP-SDE or flow matching, the ConvNeXt-V2 diamonds reach highest, ViT and Swin-T sit in the middle and the ResNet-18 circles sit lowest, near or below zero, as their rows of \cref{fig:interaction} show. On CIFAR-10, the four backbones in these columns sit, on average, in the reverse order of their real-only accuracy, with ResNet-18 lowest. On Fashion-MNIST, ResNet-18, again the most accurate, sits lowest on average in every column. On MNIST every real-only classifier is almost perfect, so every backbone has little room to gain, and all points stay close to zero. On DermaMNIST, where the training error in \cref{eq:tv-stab-dep} is largest, the order reverses, and ResNet-18 sits highest in the VP-SDE and flow-matching columns while the other backbones mostly sit below zero. On BloodMNIST, the VP-SDE and flow-matching columns again follow the reverse of real-only accuracy, with ResNet-18 lowest and ConvNeXt-V2 highest on average.}

\section{Conclusion}
\label{sec:conclusion}

\leavevmode{\marknew We asked when and why synthetic images raise or lower the test accuracy of deep classifiers when the generators see only the classifiers' own training data. Our results suggest that synthetic data help most when our classifiers detect only a small discrepancy between the generator and the real data and the classifier has room to improve, and that they help little or hurt when our classifiers detect a large discrepancy. Our excess-risk bound highlights a trade-off between the potential cost of distribution discrepancy and the benefit of a larger training set, and it limits the gain to the excess risk of the real-only classifier.}

\leavevmode{\marknew Two findings agree with the theoretical analysis. First, generators with a larger estimated lower bound on the shift tend to help less on every dataset except Fashion-MNIST, consistent with the distribution-discrepancy term in the bound. Second, on the CIFAR datasets the most accurate backbone gains the least, consistent with this limited room to gain.}

\leavevmode{\marknew We use the bound to interpret these trends rather than to predict the magnitude or direction of accuracy changes. Traditional augmentation reaches most of the largest gains at the same number of optimizer steps, so studies should compare synthetic data with traditional augmentation at the same training budget. Future work should estimate the stability and dependence terms of \cref{eq:tv-stab-dep}, which remain unquantified in our analysis, and develop more informative measures of distribution discrepancy beyond the estimated lower-bound quantity used here.

\clearpage

{
    \small
    \bibliographystyle{ieeenat_unsrt}
    \bibliography{main}
}
\clearpage

\onecolumn
\showchangesfalse
\begin{center}
 
{\Large \textbf{\thetitle}\\[0.5em]Supplementary Material}
\end{center}
\vspace{0.5em}
\appendix
\crefalias{section}{appendix}\crefalias{subsection}{subappendix}
\setcounter{table}{0}\renewcommand{\thetable}{A\arabic{table}}\renewcommand{\theHtable}{A\arabic{table}}
\setcounter{figure}{0}\renewcommand{\thefigure}{A\arabic{figure}}\renewcommand{\theHfigure}{A\arabic{figure}}

{\marknew This supplement follows the order of the main text. \Cref{app:results,app:float-rho} give the test accuracies of \cref{fig:interaction,fig:rho-blocks} and the estimated lower bounds of \cref{sec:experiments}. \Cref{app:caveats,app:float-samples,app:impl,app:float-settings,app:generators,app:float-pools} support \cref{sec:framework} with the caveats, samples, training details, settings, generator losses and FID. \Cref{app:notation,app:tv} give the notation and derivations of \cref{sec:tv}.}

\section{Full test-accuracy grids}
\label{app:results}
\begin{table}[h!]
\caption{\textbf{Test accuracy (\%) on CIFAR-10 and MNIST at $\rho = 1$.} Three-seed means, largest standard deviation 1.55 (CIFAR-10) and 0.17 (MNIST). Columns: mixing ratio $\alpha$ ($0$: real-only) and FID. Trad.\ aug.: traditional augmentation. Bold: best per backbone and dataset.}
\label{tab:main}
\begin{center}
\scriptsize
\renewcommand{\arraystretch}{0.92}
\setlength{\tabcolsep}{2.1pt}
\resizebox{\textwidth}{!}{%
\begin{tabular}{llcccccccccc|cccccccccc}
\toprule
 & & \multicolumn{10}{c|}{CIFAR-10} & \multicolumn{10}{c}{MNIST} \\
\cmidrule(lr){3-12} \cmidrule(lr){13-22}
Classifier & Generator & 0 & 0.2 & 0.4 & 0.6 & 0.8 & 1.0 & 2.0 & 4.0 & 8.0 & FID & 0 & 0.2 & 0.4 & 0.6 & 0.8 & 1.0 & 2.0 & 4.0 & 8.0 & FID\\
\midrule
ViT & VAE & 85.44 & 85.14 & 85.49 & 85.86 & 86.18 & 85.50 & 87.39 & 87.51 & 88.07 & 113.9 & 99.01 & 98.99 & 99.03 & 98.99 & 99.09 & 99.05 & 99.06 & 99.19 & 99.14 & 20.9 \\
 & DDPM & 85.44 & 86.78 & 87.41 & 87.93 & 88.83 & 88.82 & 90.12 & 91.96 & \textbf{93.31} & 11.0 & 99.01 & 98.88 & 98.97 & 98.91 & 98.90 & 98.86 & 98.85 & 98.92 & 98.94 & 5.0 \\
 & VP-SDE & 85.44 & 86.38 & 87.59 & 88.24 & 88.63 & 89.09 & 90.50 & 91.54 & 92.73 & 6.7 & 99.01 & 98.99 & 99.07 & 99.04 & 99.09 & 99.11 & 99.08 & 99.21 & 99.21 & 8.0 \\
 & Flow matching & 85.44 & 86.63 & 87.34 & 88.04 & 88.84 & 88.74 & 90.24 & 91.71 & 92.60 & 9.3 & 99.01 & 99.02 & 99.00 & 98.92 & 99.02 & 99.06 & 99.04 & 99.11 & 99.15 & 3.0 \\
 & Trad.\ aug. & 85.44 & 86.27 & 86.49 & 86.85 & 87.18 & 87.60 & 89.39 & 90.46 & 91.89 & 37.5 & 99.01 & 99.23 & 99.28 & 99.42 & 99.46 & 99.42 & 99.49 & 99.60 & \textbf{99.66} & 8.9 \\
\midrule
Swin-T & VAE & 87.67 & 87.38 & 87.73 & 87.69 & 87.84 & 87.64 & 87.88 & 88.25 & 87.99 & 113.9 & 99.16 & 99.10 & 99.04 & 99.16 & 99.12 & 99.02 & 99.06 & 99.10 & 99.18 & 20.9 \\
 & DDPM & 87.67 & 88.39 & 89.08 & 89.83 & 89.97 & 90.56 & 91.57 & 92.54 & \textbf{93.41} & 11.0 & 99.16 & 99.13 & 99.03 & 99.12 & 98.92 & 98.94 & 99.04 & 99.05 & 98.93 & 5.0 \\
 & VP-SDE & 87.67 & 88.71 & 89.30 & 89.55 & 90.26 & 90.51 & 91.38 & 92.45 & 93.12 & 6.7 & 99.16 & 99.13 & 99.14 & 99.12 & 99.12 & 99.06 & 99.00 & 99.00 & 98.98 & 8.0 \\
 & Flow matching & 87.67 & 88.43 & 88.99 & 89.52 & 90.10 & 90.26 & 91.45 & 92.32 & 93.08 & 9.3 & 99.16 & 99.16 & 99.11 & 99.15 & 99.16 & 99.18 & 99.13 & 99.18 & 99.22 & 3.0 \\
 & Trad.\ aug. & 87.67 & 88.26 & 88.67 & 88.95 & 89.27 & 89.73 & 90.76 & 91.82 & 92.47 & 37.5 & 99.16 & 99.27 & 99.34 & 99.34 & 99.42 & 99.41 & 99.49 & 99.56 & \textbf{99.57} & 8.9 \\
\midrule
ConvNeXt-V2 & VAE & 82.52 & 82.01 & 82.15 & 83.03 & 83.09 & 82.55 & 80.88 & 81.97 & 83.54 & 113.9 & 99.17 & 99.18 & 99.13 & 99.12 & 99.15 & 99.02 & 99.00 & 99.03 & 99.18 & 20.9 \\
 & DDPM & 82.52 & 83.66 & 84.16 & 85.36 & 85.96 & 86.32 & 87.68 & 90.37 & \textbf{91.94} & 11.0 & 99.17 & 99.08 & 99.10 & 99.04 & 98.97 & 98.97 & 98.82 & 98.93 & 99.01 & 5.0 \\
 & VP-SDE & 82.52 & 83.86 & 84.78 & 86.13 & 86.08 & 86.76 & 87.98 & 90.21 & \textbf{91.94} & 6.7 & 99.17 & 99.10 & 99.09 & 99.03 & 99.02 & 98.95 & 99.06 & 98.87 & 98.98 & 8.0 \\
 & Flow matching & 82.52 & 84.00 & 84.55 & 85.91 & 85.91 & 86.90 & 88.27 & 90.23 & 91.69 & 9.3 & 99.17 & 99.15 & 99.22 & 99.17 & 99.12 & 99.13 & 99.04 & 99.14 & 99.08 & 3.0 \\
 & Trad.\ aug. & 82.52 & 83.32 & 83.69 & 84.53 & 84.73 & 85.77 & 88.09 & 89.91 & 90.70 & 37.5 & 99.17 & 99.43 & 99.44 & 99.40 & 99.42 & 99.55 & 99.51 & 99.54 & \textbf{99.68} & 8.9 \\
\midrule
ResNet-18 & VAE & 94.85 & 94.22 & 94.05 & 94.00 & 93.73 & 93.38 & 92.97 & 92.40 & 91.42 & 113.9 & 99.58 & 99.64 & 99.62 & 99.64 & 99.61 & 99.64 & 99.63 & 99.53 & 99.60 & 20.9 \\
 & DDPM & 94.85 & 94.48 & 94.36 & 94.52 & 94.63 & 94.51 & 94.58 & 94.43 & 94.74 & 11.0 & 99.58 & 99.44 & 99.42 & 99.27 & 99.24 & 99.25 & 99.09 & 99.07 & 98.99 & 5.0 \\
 & VP-SDE & 94.85 & 94.61 & 94.65 & 94.52 & 94.64 & 94.44 & 94.56 & 94.26 & 94.43 & 6.7 & 99.58 & 99.54 & 99.52 & 99.49 & 99.41 & 99.43 & 99.32 & 99.19 & 99.10 & 8.0 \\
 & Flow matching & 94.85 & 94.50 & 94.47 & 94.51 & 94.31 & 94.31 & 94.24 & 94.00 & 94.17 & 9.3 & 99.58 & 99.51 & 99.52 & 99.50 & 99.48 & 99.43 & 99.37 & 99.35 & 99.27 & 3.0 \\
 & Trad.\ aug. & 94.85 & 95.03 & 94.89 & 94.98 & 95.12 & 95.11 & 95.06 & 95.61 & \textbf{95.69} & 37.5 & 99.58 & 99.62 & 99.66 & 99.67 & \textbf{99.68} & 99.67 & 99.64 & 99.66 & 99.67 & 8.9 \\
\bottomrule
\end{tabular}}
\end{center}
\end{table}
\begin{table}[h!]
\caption{\textbf{Test accuracy (\%) on CIFAR-100 and Fashion-MNIST at $\rho = 1$.} Layout as in \cref{tab:main}. Largest standard deviation 1.66 (CIFAR-100) and 0.64 (Fashion-MNIST).}
\label{tab:main2}
\begin{center}
\scriptsize
\renewcommand{\arraystretch}{0.92}
\setlength{\tabcolsep}{2.1pt}
\resizebox{\textwidth}{!}{%
\begin{tabular}{llcccccccccc|cccccccccc}
\toprule
 & & \multicolumn{10}{c|}{CIFAR-100} & \multicolumn{10}{c}{Fashion-MNIST} \\
\cmidrule(lr){3-12} \cmidrule(lr){13-22}
Classifier & Generator & 0 & 0.2 & 0.4 & 0.6 & 0.8 & 1.0 & 2.0 & 4.0 & 8.0 & FID & 0 & 0.2 & 0.4 & 0.6 & 0.8 & 1.0 & 2.0 & 4.0 & 8.0 & FID\\
\midrule
ViT & VAE & 58.69 & 58.13 & 58.75 & 58.39 & 58.70 & 58.97 & 61.18 & 63.27 & 63.80 & 101.2 & 93.72 & 93.60 & 93.84 & 93.71 & 93.94 & 93.77 & 93.74 & 93.84 & 93.77 & 47.4 \\
 & DDPM & 58.69 & 59.84 & 61.29 & 61.18 & 61.79 & 63.06 & 65.53 & 69.62 & \textbf{73.09} & 9.4 & 93.72 & 93.68 & 93.87 & 93.88 & 93.61 & 93.82 & 93.87 & 93.79 & 93.65 & 4.3 \\
 & VP-SDE & 58.69 & 59.85 & 61.42 & 61.46 & 62.74 & 63.07 & 66.37 & 69.09 & 72.02 & 6.3 & 93.72 & 93.80 & 93.87 & 93.89 & 94.03 & 94.02 & 93.85 & 94.22 & 94.10 & 5.8 \\
 & Flow matching & 58.69 & 59.79 & 61.01 & 61.17 & 61.54 & 62.40 & 65.23 & 68.13 & 71.31 & 11.0 & 93.72 & 93.78 & 93.76 & 93.80 & 93.81 & 93.76 & 93.74 & 93.73 & 93.47 & 5.0 \\
 & Trad.\ aug. & 58.69 & 59.10 & 60.60 & 60.41 & 60.70 & 61.29 & 62.65 & 65.42 & 68.55 & 39.7 & 93.72 & 93.65 & 93.88 & 93.88 & 94.01 & 94.08 & 94.12 & 94.22 & \textbf{94.48} & 19.1 \\
\midrule
Swin-T & VAE & 58.66 & 58.18 & 57.97 & 58.46 & 58.69 & 58.63 & 58.61 & 58.84 & 59.64 & 101.2 & 92.90 & 93.21 & 93.36 & 93.34 & 93.38 & 93.55 & 93.71 & 94.10 & 94.21 & 47.4 \\
 & DDPM & 58.66 & 60.33 & 61.42 & 62.09 & 62.88 & 63.93 & 66.92 & 69.87 & \textbf{72.21} & 9.4 & 92.90 & 92.92 & 93.21 & 93.36 & 93.54 & 93.44 & 93.72 & 93.64 & 93.80 & 4.3 \\
 & VP-SDE & 58.66 & 60.40 & 61.86 & 63.21 & 63.61 & 64.47 & 67.58 & 69.66 & 71.73 & 6.3 & 92.90 & 93.01 & 93.26 & 93.35 & 93.43 & 93.56 & 93.93 & 93.90 & 93.88 & 5.8 \\
 & Flow matching & 58.66 & 59.60 & 61.03 & 61.81 & 62.38 & 63.01 & 65.65 & 68.24 & 70.26 & 11.0 & 92.90 & 92.90 & 93.14 & 93.44 & 93.50 & 93.79 & 93.78 & 93.79 & 93.85 & 5.0 \\
 & Trad.\ aug. & 58.66 & 59.11 & 59.73 & 60.92 & 61.08 & 61.41 & 63.29 & 65.45 & 67.75 & 39.7 & 92.90 & 93.14 & 93.38 & 93.65 & 93.69 & 93.88 & 94.06 & 94.16 & \textbf{94.31} & 19.1 \\
\midrule
ConvNeXt-V2 & VAE & 41.92 & 41.01 & 40.37 & 41.84 & 41.53 & 42.48 & 44.97 & 47.42 & 50.85 & 101.2 & 92.47 & 92.94 & 93.29 & 93.21 & 93.34 & 93.42 & 93.21 & 93.65 & 93.59 & 47.4 \\
 & DDPM & 41.92 & 42.99 & 44.35 & 45.35 & 46.76 & 47.32 & 53.82 & 59.94 & 65.71 & 9.4 & 92.47 & 92.52 & 92.76 & 92.70 & 92.76 & 92.42 & 92.81 & 93.13 & 93.55 & 4.3 \\
 & VP-SDE & 41.92 & 44.29 & 46.16 & 48.25 & 49.89 & 51.43 & 56.04 & 62.81 & \textbf{66.41} & 6.3 & 92.47 & 92.55 & 92.73 & 92.81 & 92.73 & 92.94 & 93.15 & 93.37 & 93.55 & 5.8 \\
 & Flow matching & 41.92 & 43.58 & 45.71 & 46.26 & 48.16 & 49.60 & 54.58 & 60.73 & 63.76 & 11.0 & 92.47 & 92.72 & 92.63 & 92.74 & 92.76 & 93.07 & 92.96 & 93.12 & \textbf{93.67} & 5.0 \\
 & Trad.\ aug. & 41.92 & 42.73 & 42.88 & 45.42 & 46.18 & 46.55 & 52.19 & 55.12 & 59.52 & 39.7 & 92.47 & 92.53 & 92.59 & 92.72 & 92.97 & 92.82 & 93.26 & 93.51 & 93.61 & 19.1 \\
\midrule
ResNet-18 & VAE & 76.66 & 74.60 & 74.00 & 73.54 & 72.65 & 72.62 & 71.12 & 69.50 & 67.90 & 101.2 & 95.07 & 95.00 & 95.02 & 95.07 & 95.07 & 94.78 & 94.88 & 94.84 & 94.76 & 47.4 \\
 & DDPM & 76.66 & 76.08 & 76.16 & 76.00 & 76.10 & 75.93 & 76.41 & 76.51 & 77.11 & 9.4 & 95.07 & 94.79 & 94.90 & 94.53 & 94.56 & 94.65 & 94.55 & 94.31 & 94.24 & 4.3 \\
 & VP-SDE & 76.66 & 76.00 & 76.36 & 76.24 & 75.95 & 75.85 & 76.08 & 76.10 & 76.25 & 6.3 & 95.07 & 95.04 & 94.93 & 94.82 & 94.82 & 95.02 & 94.78 & 94.70 & 94.62 & 5.8 \\
 & Flow matching & 76.66 & 76.03 & 75.45 & 75.23 & 75.12 & 74.92 & 74.76 & 75.03 & 75.23 & 11.0 & 95.07 & \textbf{95.11} & 94.86 & 94.84 & 94.62 & 94.75 & 94.68 & 94.53 & 94.34 & 5.0 \\
 & Trad.\ aug. & 76.66 & 76.47 & 76.43 & 76.68 & 76.78 & 76.85 & 77.50 & 77.96 & \textbf{78.21} & 39.7 & 95.07 & 95.09 & 95.07 & 95.00 & 95.10 & 95.04 & 95.08 & 94.97 & 94.99 & 19.1 \\
\bottomrule
\end{tabular}}
\end{center}
\end{table}
\begin{table}[h!]
\caption{\textbf{Test accuracy (\%) on DermaMNIST and BloodMNIST at $\rho = 1$.} Layout as in \cref{tab:main}. Largest standard deviation 1.80 (DermaMNIST) and 0.55 (BloodMNIST).}
\label{tab:main3}
\begin{center}
\scriptsize
\renewcommand{\arraystretch}{0.92}
\setlength{\tabcolsep}{2.1pt}
\resizebox{\textwidth}{!}{%
\begin{tabular}{llcccccccccc|cccccccccc}
\toprule
 & & \multicolumn{10}{c|}{DermaMNIST} & \multicolumn{10}{c}{BloodMNIST} \\
\cmidrule(lr){3-12} \cmidrule(lr){13-22}
Classifier & Generator & 0 & 0.2 & 0.4 & 0.6 & 0.8 & 1.0 & 2.0 & 4.0 & 8.0 & FID & 0 & 0.2 & 0.4 & 0.6 & 0.8 & 1.0 & 2.0 & 4.0 & 8.0 & FID\\
\midrule
ViT & VAE & 77.64 & 76.43 & 76.92 & 76.79 & 77.52 & 77.92 & 77.42 & \textbf{78.50} & 77.97 & 54.2 & 97.83 & 97.87 & 97.97 & 97.90 & 97.79 & 97.85 & 98.06 & 97.95 & 97.88 & 127.3 \\
 & DDPM & 77.64 & 76.41 & 75.33 & 75.46 & 74.91 & 75.34 & 74.90 & 74.88 & 72.62 & 33.3 & 97.83 & 97.88 & 97.85 & 97.81 & 97.63 & 98.10 & 97.79 & 97.82 & 97.91 & 5.3 \\
 & VP-SDE & 77.64 & 76.97 & 76.69 & 76.21 & 76.77 & 76.31 & 75.71 & 76.04 & 75.66 & 25.2 & 97.83 & 97.87 & 98.10 & 98.20 & 98.16 & 98.19 & 98.30 & 98.26 & 98.34 & 4.8 \\
 & Flow matching & 77.64 & 77.69 & 77.11 & 76.82 & 76.23 & 76.44 & 75.48 & 74.78 & 75.05 & 21.8 & 97.83 & 98.06 & 98.02 & 98.19 & 98.02 & 98.19 & 98.29 & 98.35 & 98.35 & 3.4 \\
 & Trad.\ aug. & 77.64 & 76.82 & 77.67 & 77.97 & 77.01 & 78.49 & 77.24 & 77.32 & 78.40 & 132.5 & 97.83 & 97.96 & 97.87 & 98.01 & 98.00 & 98.16 & 98.36 & 97.91 & \textbf{98.50} & 110.9 \\
\midrule
Swin-T & VAE & 77.34 & 76.63 & 76.86 & 76.43 & 76.96 & 77.22 & 77.02 & 77.17 & 77.31 & 54.2 & 97.85 & 97.87 & 97.84 & 97.86 & 98.04 & 97.76 & 97.60 & 98.05 & 98.02 & 127.3 \\
 & DDPM & 77.34 & 75.69 & 75.63 & 75.84 & 75.26 & 75.39 & 74.05 & 73.93 & 73.67 & 33.3 & 97.85 & 98.03 & 97.63 & 98.01 & 98.01 & 98.13 & 98.20 & 98.02 & 98.12 & 5.3 \\
 & VP-SDE & 77.34 & 76.34 & 76.43 & 74.95 & 75.94 & 75.86 & 76.48 & 74.60 & 75.43 & 25.2 & 97.85 & 97.94 & 98.13 & 98.15 & 98.34 & 98.35 & 98.30 & 98.56 & 98.54 & 4.8 \\
 & Flow matching & 77.34 & 77.07 & 76.54 & 76.48 & 75.05 & 76.03 & 75.83 & 75.88 & 76.36 & 21.8 & 97.85 & 97.97 & 98.10 & 98.24 & 98.38 & 98.41 & 98.48 & 98.60 & 98.47 & 3.4 \\
 & Trad.\ aug. & 77.34 & 76.58 & 77.76 & 77.82 & 78.14 & 78.79 & 78.32 & 79.29 & \textbf{79.78} & 132.5 & 97.85 & 97.93 & 98.06 & 97.95 & 98.00 & 98.10 & 98.49 & 98.63 & \textbf{98.82} & 110.9 \\
\midrule
ConvNeXt-V2 & VAE & 74.13 & 74.43 & 73.65 & 74.01 & 74.15 & 73.55 & 73.73 & 74.08 & 73.05 & 54.2 & 96.97 & 96.98 & 96.78 & 96.85 & 96.77 & 96.73 & 96.92 & 96.67 & 97.08 & 127.3 \\
 & DDPM & 74.13 & 74.46 & 73.30 & 73.22 & 71.59 & 71.89 & 70.92 & 70.01 & 69.33 & 33.3 & 96.97 & 96.82 & 96.58 & 96.75 & 96.38 & 96.79 & 96.49 & 96.16 & 96.76 & 5.3 \\
 & VP-SDE & 74.13 & 74.86 & 73.80 & 74.38 & 72.90 & 73.18 & 72.73 & 72.04 & 72.80 & 25.2 & 96.97 & 97.12 & 97.33 & 96.86 & 97.25 & 97.29 & 97.81 & 97.66 & 98.09 & 4.8 \\
 & Flow matching & 74.13 & 74.46 & 74.75 & 74.21 & 74.13 & 74.70 & 73.38 & 72.75 & 72.34 & 21.8 & 96.97 & 97.23 & 97.33 & 97.42 & 97.20 & 97.31 & 97.76 & 97.65 & \textbf{98.19} & 3.4 \\
 & Trad.\ aug. & 74.13 & \textbf{76.01} & 75.31 & 75.99 & 75.18 & 74.36 & 75.26 & 75.01 & 74.90 & 132.5 & 96.97 & 96.62 & 97.04 & 96.93 & 97.26 & 96.95 & 97.32 & 97.45 & 97.58 & 110.9 \\
\midrule
ResNet-18 & VAE & 78.35 & 77.97 & 77.99 & 77.89 & 78.75 & 78.87 & 79.12 & 79.47 & 80.22 & 54.2 & 98.59 & 98.89 & 98.75 & 98.71 & 98.83 & 98.82 & 98.81 & 98.68 & 98.70 & 127.3 \\
 & DDPM & 78.35 & 77.49 & 77.54 & 77.22 & 76.67 & 77.29 & 77.27 & 77.01 & 77.24 & 33.3 & 98.59 & 98.81 & 98.75 & 98.73 & 98.72 & 98.66 & 98.62 & 98.33 & 98.29 & 5.3 \\
 & VP-SDE & 78.35 & 78.34 & 77.61 & 78.67 & 78.87 & 79.04 & 80.55 & 80.50 & \textbf{82.83} & 25.2 & 98.59 & 98.77 & 98.77 & 98.78 & 98.74 & 98.78 & 98.87 & 98.83 & 98.72 & 4.8 \\
 & Flow matching & 78.35 & 78.09 & 79.14 & 78.92 & 79.25 & 78.80 & 79.88 & 81.73 & 81.50 & 21.8 & 98.59 & 98.79 & 98.76 & 98.79 & 98.82 & 98.67 & 98.76 & 98.92 & 98.94 & 3.4 \\
 & Trad.\ aug. & 78.35 & 79.73 & 80.07 & 80.18 & 80.98 & 81.06 & 80.35 & 81.25 & 80.42 & 132.5 & 98.59 & 98.84 & 98.71 & 98.77 & 98.89 & 98.92 & \textbf{98.97} & 98.87 & 98.67 & 110.9 \\
\bottomrule
\end{tabular}}
\end{center}
\end{table}

\clearpage
\section{Lower real-data fractions and estimated lower bounds}
\label{app:float-rho}
\begin{table}[h!]
\caption{\textbf{Test accuracy (\%) at each real-data fraction $\rho$ on MNIST and CIFAR-10.} Mixtures at $\alpha = 1.0$, three-seed means. $\alpha{=}0$: real-only classifier at that $\rho$. FM: flow matching. Trad.: traditional augmentation.}
\label{tab:rho}
\begin{center}
\scriptsize
\setlength{\tabcolsep}{4.5pt}
\begin{tabular}{llcccccc|cccccc}
\toprule
 & & \multicolumn{6}{c|}{MNIST} & \multicolumn{6}{c}{CIFAR-10} \\
\cmidrule(lr){3-8} \cmidrule(lr){9-14}
Classifier & $\rho$ & $\alpha{=}0$ & VAE & DDPM & VP-SDE & FM & Trad. & $\alpha{=}0$ & VAE & DDPM & VP-SDE & FM & Trad. \\
\midrule
ResNet-18   & 0.2 & 99.09 & 99.01 & 98.55 & 98.67 & 98.85 & 99.32 & 84.20 & 84.74 & {\marknew 85.55} & 86.26 & 85.79 & 87.25 \\
            & 0.4 & 99.36 & 99.39 & 98.96 & 99.14 & 99.20 & 99.55 & 90.24 & 89.44 & 90.95 & 90.90 & 90.61 & 91.35 \\
            & 0.6 & 99.43 & 99.45 & 99.24 & 99.36 & 99.35 & 99.63 & 92.50 & 91.51 & {\marknew 92.57} & 92.61 & 92.62 & 93.36 \\
            & 0.8 & 99.57 & 99.60 & 99.28 & 99.29 & 99.46 & 99.63 & 93.76 & 92.78 & 93.52 & 93.54 & 93.56 & 94.24 \\
            & 1.0 & 99.58 & 99.64 & 99.25 & 99.43 & 99.43 & 99.67 & 94.85 & 93.38 & 94.51 & 94.44 & 94.31 & 95.11 \\
\midrule
ViT         & 0.2 & 97.60 & 97.37 & 97.28 & 97.50 & 97.45 & 98.48 & 69.74 & 70.87 & 73.48 & 74.97 & 73.84 & 72.45 \\
            & 0.4 & 98.44 & 98.34 & 98.28 & 98.40 & 98.45 & 99.19 & 77.44 & 77.84 & 81.56 & 81.26 & 81.04 & 78.91 \\
            & 0.6 & 98.56 & 98.69 & 98.63 & 98.80 & 98.79 & 99.21 & 81.29 & 81.37 & 85.01 & 85.02 & 84.65 & 82.99 \\
            & 0.8 & 98.92 & 98.93 & 98.89 & 98.99 & 98.90 & 99.37 & 83.84 & 83.97 & 87.29 & 87.29 & 87.46 & 85.53 \\
            & 1.0 & 99.01 & 99.05 & 98.86 & 99.11 & 99.06 & 99.42 & 85.44 & 85.50 & 88.82 & 89.09 & 88.74 & 87.60 \\
\midrule
Swin-T      & 0.2 & 98.08 & 97.86 & 98.03 & 97.73 & 97.73 & 98.80 & 68.63 & 70.54 & 74.14 & 75.71 & 74.19 & 71.46 \\
            & 0.4 & 98.82 & 98.59 & 98.34 & 98.46 & 98.70 & 99.11 & 77.34 & 78.40 & 83.28 & 83.59 & 83.76 & 81.07 \\
            & 0.6 & 98.92 & 98.82 & 98.76 & 98.77 & 98.91 & 99.27 & 81.45 & 83.08 & 86.54 & 87.34 & 86.81 & 85.34 \\
            & 0.8 & 99.05 & 99.06 & 98.95 & 98.99 & 99.02 & 99.35 & 84.95 & 86.11 & 88.78 & 89.12 & 88.91 & 87.67 \\
            & 1.0 & 99.16 & 99.02 & 98.94 & 99.06 & 99.18 & 99.41 & 87.67 & 87.64 & 90.56 & 90.51 & 90.26 & 89.73 \\
\midrule
ConvNeXt-V2 & 0.2 & 98.10 & 97.65 & 97.82 & 97.63 & 97.98 & 98.84 & 61.04 & 60.40 & 65.93 & 71.01 & 69.25 & 66.85 \\
            & 0.4 & 98.84 & 98.44 & 98.47 & 98.53 & 98.66 & 99.11 & 71.04 & 69.70 & 76.28 & 78.81 & 77.54 & 75.08 \\
            & 0.6 & 98.98 & 98.81 & 98.70 & 98.87 & 98.97 & 99.41 & 75.66 & 75.16 & 81.94 & 82.69 & 82.10 & 79.73 \\
            & 0.8 & 99.06 & 98.97 & 98.93 & 98.77 & 98.92 & 99.49 & 79.39 & 78.53 & 84.64 & 84.59 & 84.71 & 84.02 \\
            & 1.0 & 99.17 & 99.02 & 98.97 & 98.95 & 99.13 & 99.55 & 82.52 & 82.55 & 86.32 & 86.76 & 86.90 & 85.77 \\
\bottomrule
\end{tabular}
\end{center}
\end{table}
\begin{table}[h!]
\caption{\textbf{Test accuracy (\%) at each real-data fraction $\rho$ on CIFAR-100 and Fashion-MNIST.} Layout as in \cref{tab:rho}.}
\label{tab:rho2}
\begin{center}
\scriptsize
\setlength{\tabcolsep}{4.5pt}
\begin{tabular}{llcccccc|cccccc}
\toprule
 & & \multicolumn{6}{c|}{CIFAR-100} & \multicolumn{6}{c}{Fashion-MNIST} \\
\cmidrule(lr){3-8} \cmidrule(lr){9-14}
Classifier & $\rho$ & $\alpha{=}0$ & VAE & DDPM & VP-SDE & FM & Trad. & $\alpha{=}0$ & VAE & DDPM & VP-SDE & FM & Trad. \\
\midrule
ResNet-18   & 0.2 & 51.07 & 49.00 & 53.46 & 55.39 & 52.83 & 55.56 & 91.99 & 92.35 & 92.06 & 92.17 & 92.06 & 92.04 \\
            & 0.4 & 64.94 & 60.37 & 64.76 & 64.68 & 63.11 & 66.91 & 93.55 & 93.68 & 93.27 & 93.53 & 93.54 & 93.49 \\
            & 0.6 & 71.10 & 66.89 & 69.78 & 69.77 & 67.79 & 71.83 & 94.45 & 94.34 & 93.91 & 94.01 & 94.01 & 94.24 \\
            & 0.8 & 74.22 & 70.46 & 72.44 & 72.53 & 71.03 & 75.07 & 94.91 & 94.75 & 94.37 & 94.44 & 94.38 & 94.78 \\
            & 1.0 & 76.66 & 72.62 & 75.93 & 75.85 & 74.92 & 76.85 & 95.07 & 94.78 & 94.65 & 95.02 & 94.75 & 95.04 \\
\midrule
ViT         & 0.2 & 36.61 & 35.89 & 37.94 & 39.82 & 37.66 & 37.45 & 89.80 & 90.33 & 90.63 & 90.60 & 90.50 & 90.66 \\
            & 0.4 & 46.15 & 45.57 & 48.64 & 49.27 & 48.34 & 47.38 & 91.74 & 92.08 & 92.27 & 92.42 & 92.36 & 92.50 \\
            & 0.6 & 51.86 & 51.52 & 54.77 & 55.11 & 54.33 & 53.92 & 92.54 & 92.91 & 93.22 & 93.27 & 93.05 & 93.24 \\
            & 0.8 & 55.89 & 55.44 & 59.33 & 58.86 & 58.12 & 57.89 & 93.33 & 93.49 & 93.73 & 93.78 & 93.57 & 93.54 \\
            & 1.0 & 58.69 & 58.97 & 63.06 & 63.07 & 62.40 & 61.29 & 93.72 & 93.77 & 93.82 & 94.02 & 93.76 & 94.08 \\
\midrule
Swin-T      & 0.2 & 28.84 & 31.42 & 32.16 & 37.02 & 35.07 & 31.80 & 89.08 & 89.97 & 90.03 & 89.85 & 89.63 & 90.35 \\
            & 0.4 & 41.15 & 42.07 & 47.49 & 49.24 & 47.67 & 45.01 & 90.95 & 91.43 & 91.86 & 91.51 & 91.72 & 91.79 \\
            & 0.6 & 49.04 & 50.08 & 54.57 & 56.04 & 54.52 & 52.79 & 91.82 & 92.31 & 92.37 & 92.43 & 92.42 & 92.51 \\
            & 0.8 & 54.90 & 54.95 & 59.06 & 60.16 & 58.04 & 57.87 & 92.47 & 92.93 & 93.00 & 93.13 & 93.16 & 93.25 \\
            & 1.0 & 58.66 & 58.63 & 63.93 & 64.47 & 63.01 & 61.41 & 92.90 & 93.55 & 93.44 & 93.56 & 93.79 & 93.88 \\
\midrule
ConvNeXt-V2 & 0.2 & 24.57 & 23.08 & 24.76 & 26.99 & 26.54 & 24.67 & 88.83 & 89.87 & 89.55 & 89.68 & 89.60 & 88.95 \\
            & 0.4 & 30.62 & 28.33 & 33.28 & 35.49 & 34.43 & 31.51 & 90.42 & 91.33 & 91.04 & 90.99 & 91.41 & 90.56 \\
            & 0.6 & 34.81 & 33.78 & 39.01 & 41.52 & 39.91 & 38.02 & 91.46 & 92.06 & 91.70 & 91.85 & 91.90 & 91.54 \\
            & 0.8 & 38.84 & 39.89 & 42.94 & 46.91 & 45.54 & 42.83 & 92.15 & 92.84 & 92.22 & 92.42 & 92.02 & 92.46 \\
            & 1.0 & 41.92 & 42.48 & 47.32 & 51.43 & 49.60 & 46.55 & 92.47 & 93.42 & 92.42 & 92.94 & 93.07 & 92.82 \\
\bottomrule
\end{tabular}
\end{center}
\end{table}

\clearpage
\paragraph{Estimated lower bounds.}
\Cref{tab:dtv} gives $\widehat{L}(G_{\rho}) = \max_{f} | \widehat{R}^{\mathrm{test}}_{P}(f) - \widehat{R}^{\mathrm{test}}_{G_{\rho}}(f) |$ of \cref{sec:results-read}, the maximum over the $J = 36$ seed-0 classifiers trained on real data alone or with $G_{\rho}$ (four backbones, each at $\alpha \in \{0, 0.2, 0.4, 0.6, 0.8, 1.0, 2.0, 4.0, 8.0\}$). Each test error uses the $m_{P}$ test images (\cref{tab:splits}) or $m_{G} = 5{,}000$ fresh draws from $G_{\rho}$.

\begin{table}[h!]
\centering
\small
\caption{\textbf{Estimated lower bounds $\widehat{L}(G_{\rho})$ on $d_{\mathrm{TV}}(P, G_{\rho})$ at $\rho = 1$.} From 36 seed-0 classifiers per generator (4 backbones $\times$ 9 values of $\alpha$) and about 5{,}000 fresh draws from $G_{\rho}$.}
\label{tab:dtv}
\begin{tabular}{@{}lcccc@{}}
\toprule
Dataset & VAE & DDPM & VP-SDE & FM \\
\midrule
MNIST & 0.03 & 0.13 & 0.14 & 0.03 \\
Fashion-MNIST & 0.12 & 0.10 & 0.15 & 0.06 \\
CIFAR-10 & 0.77 & 0.10 & 0.03 & 0.06 \\
CIFAR-100 & 0.70 & 0.17 & 0.08 & 0.16 \\
DermaMNIST & 0.40 & 0.40 & 0.23 & 0.21 \\
BloodMNIST & 0.12 & 0.05 & 0.01 & 0.01 \\
\bottomrule
\end{tabular}
\end{table}

\begin{table}[h!]
\centering
\small
\caption{\textbf{With less real data, the estimated lower bound on $d_{\mathrm{TV}}(P, G_{\rho})$ rises for the higher-fidelity generators, most on MNIST and CIFAR-100.} The classifiers of \cref{tab:dtv}, trained at $\rho = 1$, on 5{,}000 fresh draws from $G_{\rho}$ refitted at each $\rho$. Rows at $\rho = 1$ repeat \cref{tab:dtv}.}
\label{tab:dtv-rho}
\begin{tabular}{@{}llcccc@{}}
\toprule
Dataset & $\rho$ & VAE & DDPM & VP-SDE & FM \\
\midrule
MNIST & $0.2$ & 0.06 & 0.72 & 0.52 & 0.31 \\
 & $0.4$ & 0.06 & 0.28 & 0.18 & 0.11 \\
 & $0.6$ & 0.05 & 0.19 & 0.15 & 0.04 \\
 & $0.8$ & 0.03 & 0.16 & 0.14 & 0.04 \\
 & $1.0$ & 0.03 & 0.13 & 0.14 & 0.03 \\
\midrule
Fashion-MNIST & $0.2$ & 0.16 & 0.20 & 0.18 & 0.14 \\
 & $0.4$ & 0.13 & 0.14 & 0.19 & 0.10 \\
 & $0.6$ & 0.14 & 0.12 & 0.14 & 0.08 \\
 & $0.8$ & 0.12 & 0.11 & 0.14 & 0.07 \\
 & $1.0$ & 0.12 & 0.10 & 0.15 & 0.06 \\
\midrule
CIFAR-10 & $0.2$ & 0.79 & 0.28 & 0.13 & 0.17 \\
 & $0.4$ & 0.78 & 0.15 & 0.07 & 0.11 \\
 & $0.6$ & 0.78 & 0.12 & 0.05 & 0.08 \\
 & $0.8$ & 0.78 & 0.10 & 0.03 & 0.07 \\
 & $1.0$ & 0.77 & 0.10 & 0.03 & 0.06 \\
\midrule
CIFAR-100 & $0.2$ & 0.72 & 0.50 & 0.31 & 0.43 \\
 & $0.4$ & 0.71 & 0.36 & 0.23 & 0.35 \\
 & $0.6$ & 0.71 & 0.30 & 0.20 & 0.30 \\
 & $0.8$ & 0.72 & 0.27 & 0.16 & 0.26 \\
 & $1.0$ & 0.70 & 0.17 & 0.08 & 0.16 \\
\bottomrule
\end{tabular}
\end{table}

\FloatBarrier
\section{Scope and caveats}
\label{app:caveats}
The caveats below limit how to read individual numbers.

\paragraph{Couplings within the design.}
Four couplings remain inside the design. First, we fix the classifier's training budget in epochs, so raising $\alpha$ also raises the number of optimizer steps.
{\marknew Second, we train and sample each generator once per dataset and real-data fraction, so our results cannot separate the run-to-run variance of a generator from the differences between generators.} Third, lowering $\rho$ lowers synthetic-data quality as well as the amount of real data. We refit the generators on each subsample at the same epoch count, so their image-update budget also scales with $\rho$. Fourth, we draw every synthetic dataset with equal per-class counts, so on an imbalanced dataset a mixture also flattens the training class prior while the test prior stays unchanged. On DermaMNIST the majority class holds $67.0$ percent of the real training set and $66.9$ percent of the test set, but its share of a mixture falls toward $1/7$ as $\alpha$ grows. {\marknew The four benchmark datasets are class-balanced or nearly so (the largest MNIST class holds $0.112$ of $\mathcal{D}_{\mathrm{train}}$), so the shift is negligible on them,} but on DermaMNIST we cannot separate this prior shift from synthetic-data fidelity, and BloodMNIST, with its $2.7$ to $1$ class imbalance, carries a milder version of it.

\paragraph{Stratified synthetic draws.}
Our synthetic datasets have fixed per-class counts, so given $\mathcal{D}_{\mathrm{train}}(\rho)$ and the generator's training randomness their images are independent draws from the class-conditional distributions of $G_{\rho}$, whose class prior equals the shares of these counts, not i.i.d.\ draws from $G_{\rho}$ as in \citet[Section~2.2]{zheng2023toward}. \Cref{eq:tv-stab-dep} still holds. The proof of their Theorem~3.1 uses the synthetic draws only through two moment inequalities that need independent, not identically distributed, draws \citep[Lemma~2 and Theorem~4]{bousquet2020sharper}, once each synthetic point is centered by its risk under its own class-conditional distribution. These risks sum to $\alpha n\, R_{G_{\rho}}(f)$ for every classifier $f$, because the class prior of $G_{\rho}$ equals the shares of the counts, and $\mathcal{T}$ is taken over the law of these draws.

\paragraph{Shared measurement.}
The estimated lower bounds of \cref{tab:dtv} and the gains in test accuracy of \cref{fig:cost} share the test errors of some classifiers, so part of their relation in \cref{sec:results-generator} may reflect this shared measurement rather than the generators.

\paragraph{Algorithmic priors.}
By design, no generator or backbone under study starts from pretrained weights, so these results do not say what a pretrained generator adds. No information about the target distribution beyond the training set reaches the classifier through the synthetic data. Yet the generator's architecture and objective are an image prior we chose, and we cannot separate the value of restating the training set from the value of passing it through that prior.

\paragraph{Generator quality.} At our fixed budget, the shared U-Net trails a separately tuned generator, with an FID of 11.0 for our CIFAR-10 DDPM against roughly 3.2 for a tuned one \citep{ho2020denoising}.

\paragraph{Class shares.} Our synthetic data is class-balanced ($1/K$ per class, with $K$ classes), so class imbalance alone moves $G$ away from $P$. The largest class share in $\mathcal{D}_{\mathrm{train}}$ is $0.100$ on Fashion-MNIST and CIFAR-10 ($K = 10$) and $0.010$ on CIFAR-100 ($K = 100$), which are balanced. The largest share is $0.112$ on MNIST ($K = 10$), which is nearly balanced. The largest share is $0.195$ against $1/8 = 0.125$ on BloodMNIST, which is mildly imbalanced, and $0.670$ against $1/7 \approx 0.143$ on DermaMNIST, which is strongly imbalanced. On DermaMNIST, the added images are class-balanced while the real test images are not, a label shift that alone puts every generator and traditional augmentation more than $0.5$ from $P$ in total variation, yet traditional augmentation gains while DDPM loses the most, so this shift does not explain the losses.

\clearpage
\section{Samples from every generator}
\label{app:float-samples}
\newcommand{\bloodsamplerow}{}
\IfFileExists{figures/bloodmnist_vae.pdf}{\bloodsamplestrue}{}%
\ifbloodsamples
\renewcommand{\bloodsamplerow}{%
\parbox[b][0.16\textwidth][c]{1.1em}{\centering\rotatebox[origin=c]{90}{\small BloodMNIST}} &
\includegraphics[width=0.16\textwidth]{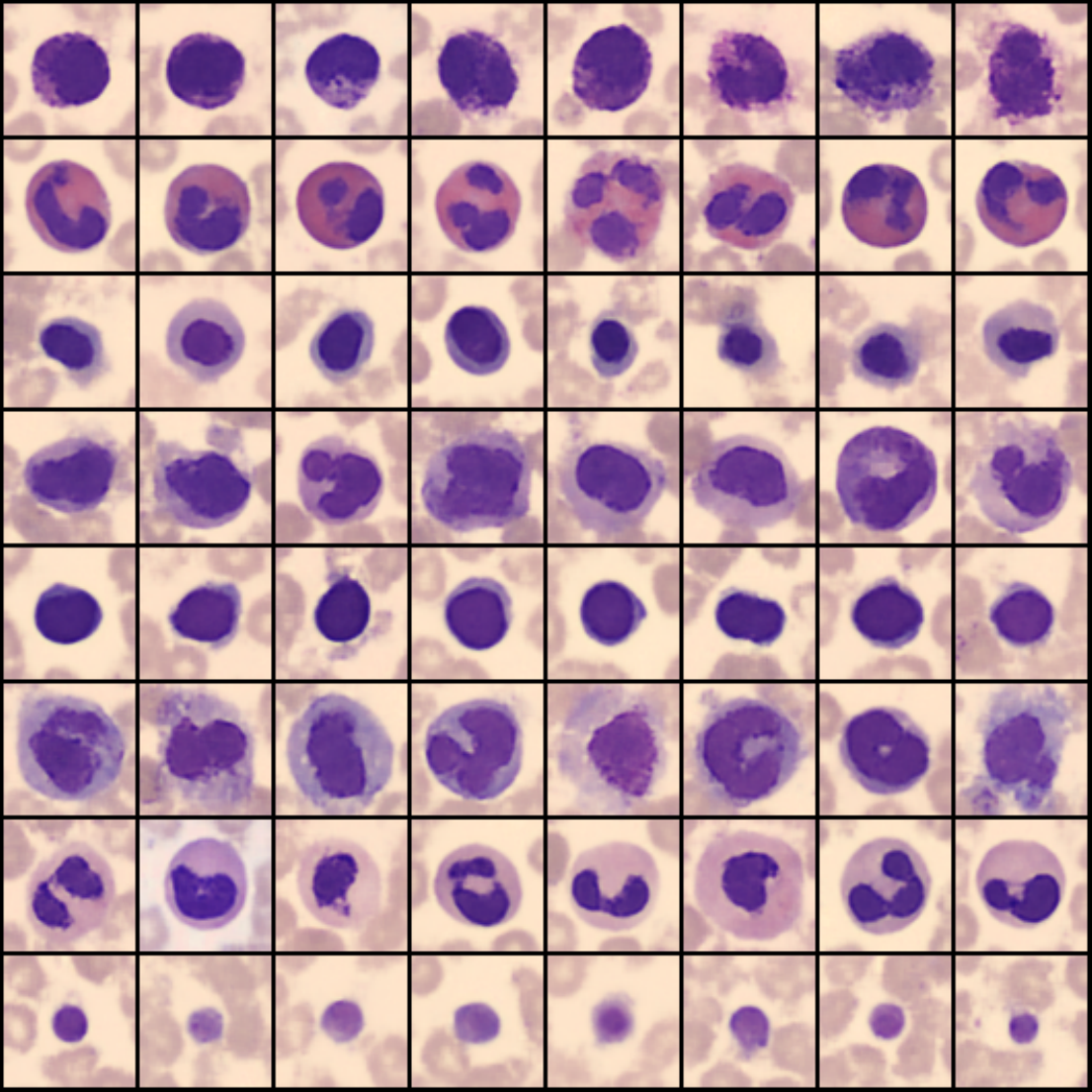} &
\includegraphics[width=0.16\textwidth]{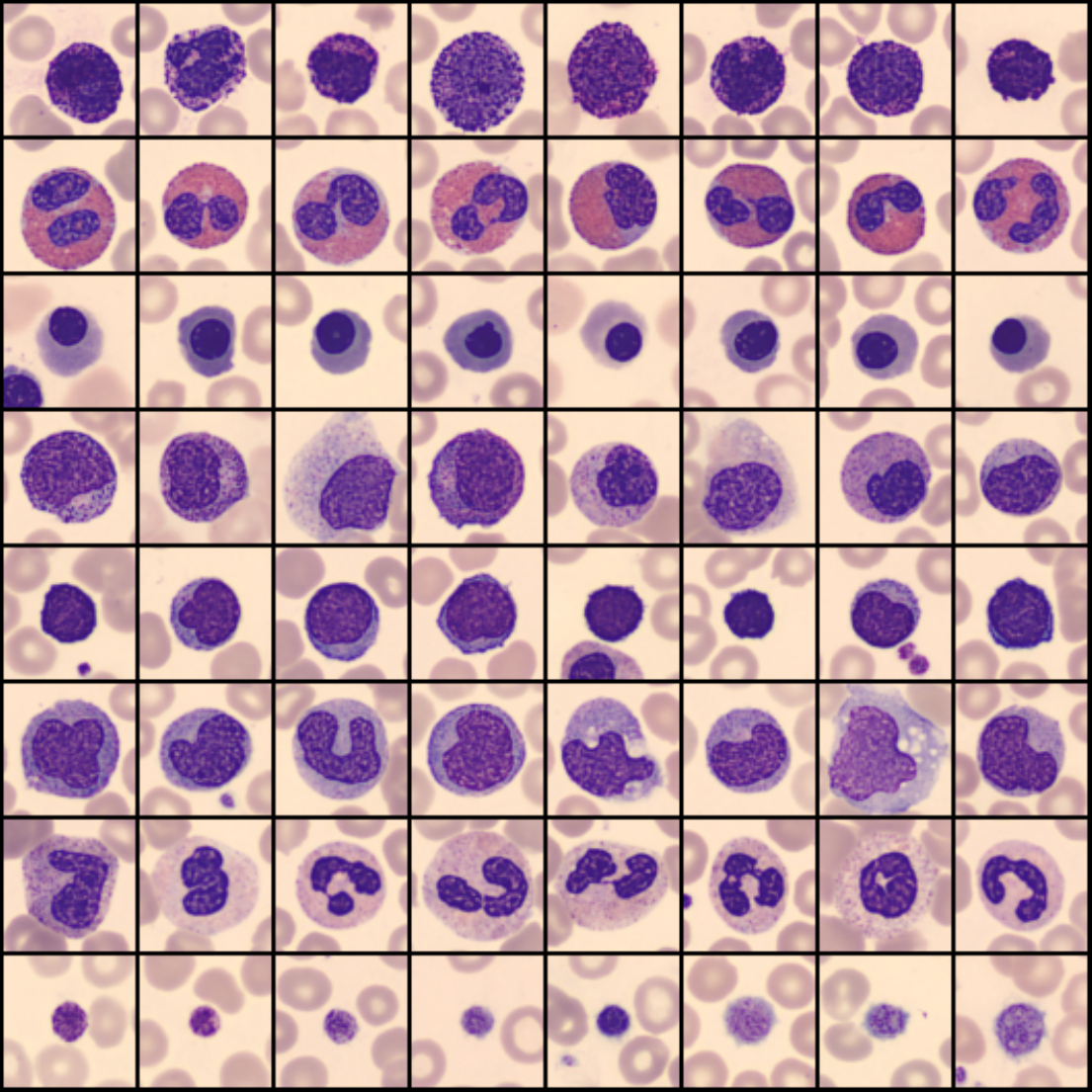} &
\includegraphics[width=0.16\textwidth]{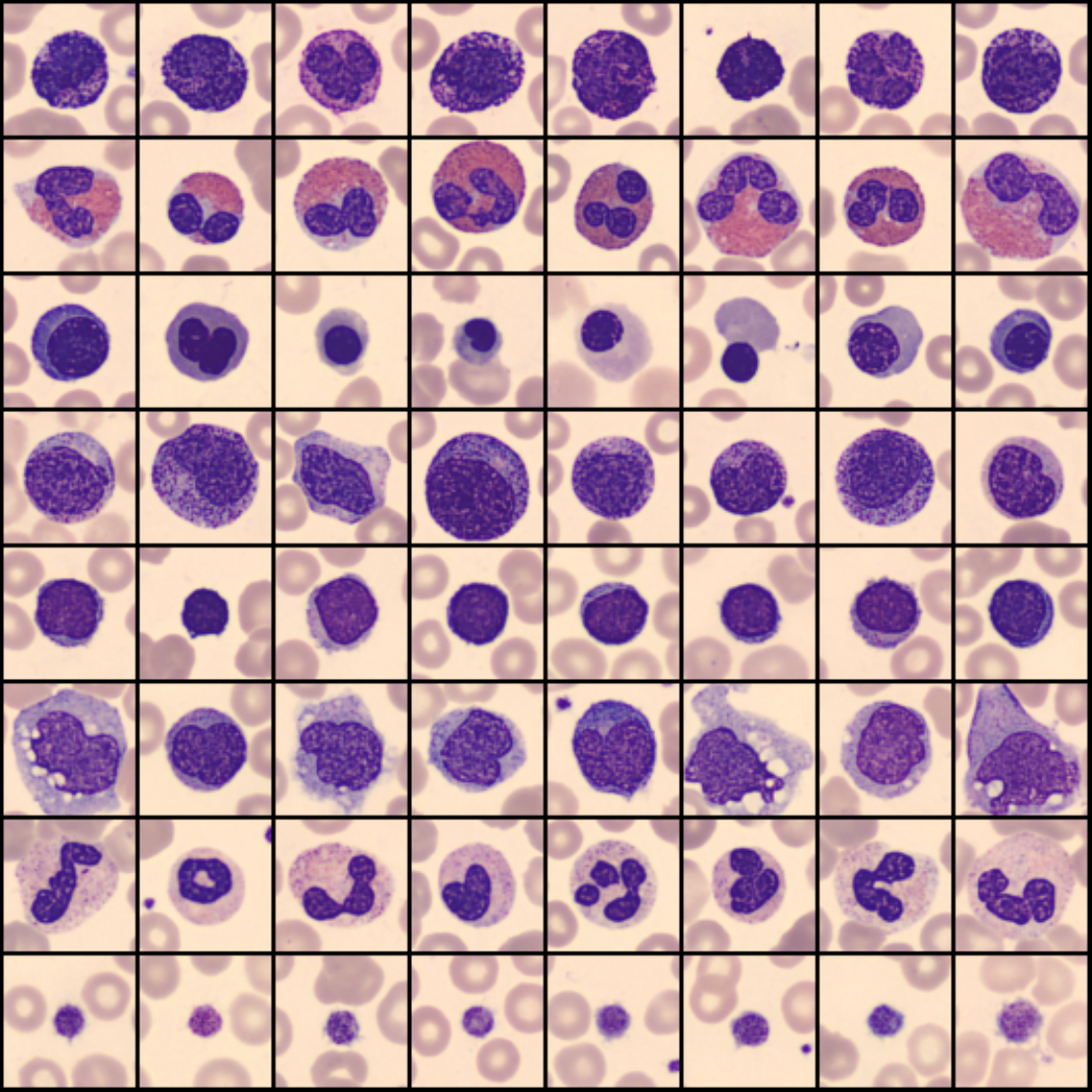} &
\includegraphics[width=0.16\textwidth]{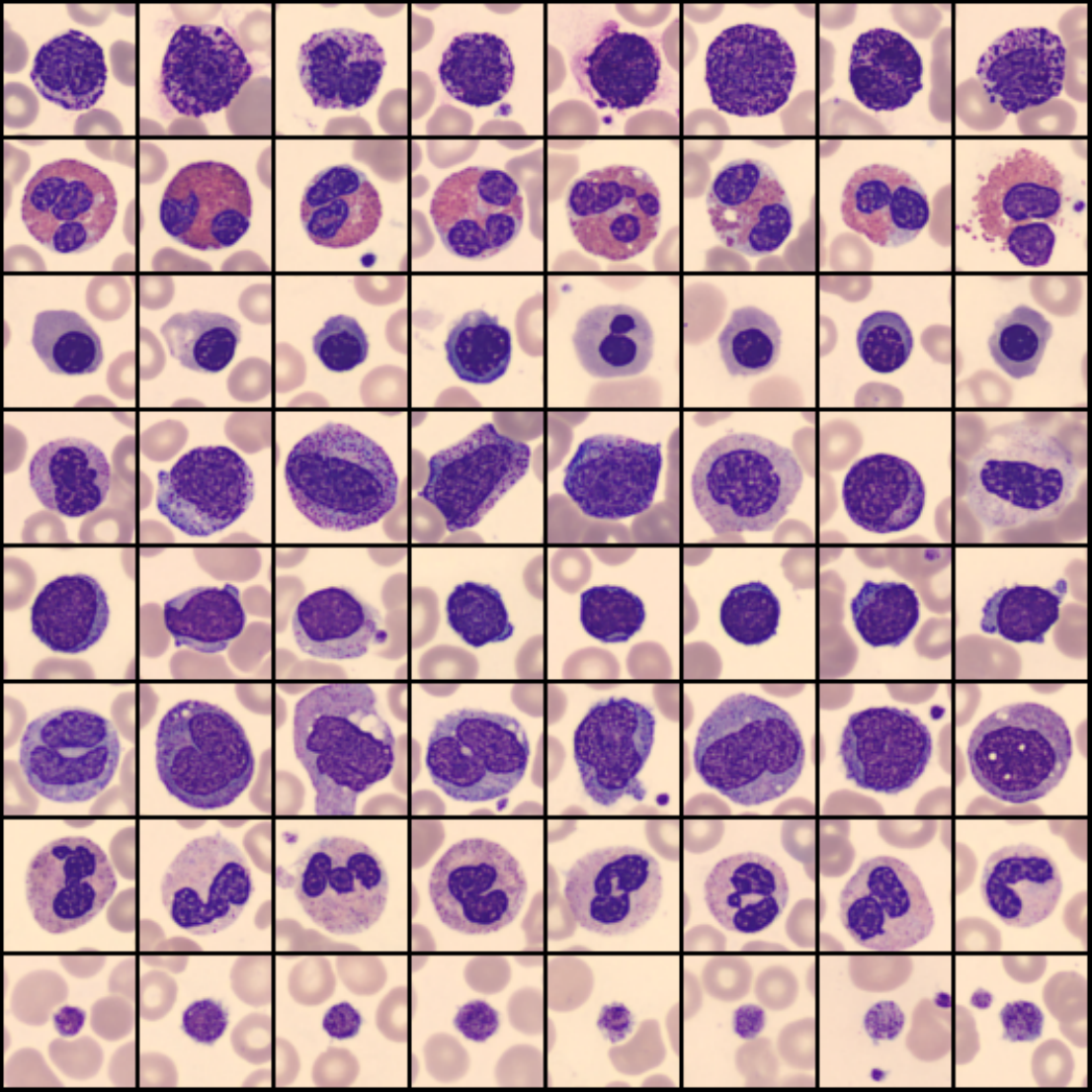} \\[2pt]}
\fi
\begin{figure}[h!]
\centering
\setlength{\tabcolsep}{1.5pt}
\renewcommand{\arraystretch}{1.0}
\begin{tabular}{@{}c@{\hspace{3pt}}cccc@{}}
 & \small VAE & \small DDPM & \small VP-SDE & \small Flow matching \\
\parbox[b][0.16\textwidth][c]{1.1em}{\centering\rotatebox[origin=c]{90}{\small DermaMNIST}} &
\includegraphics[width=0.16\textwidth]{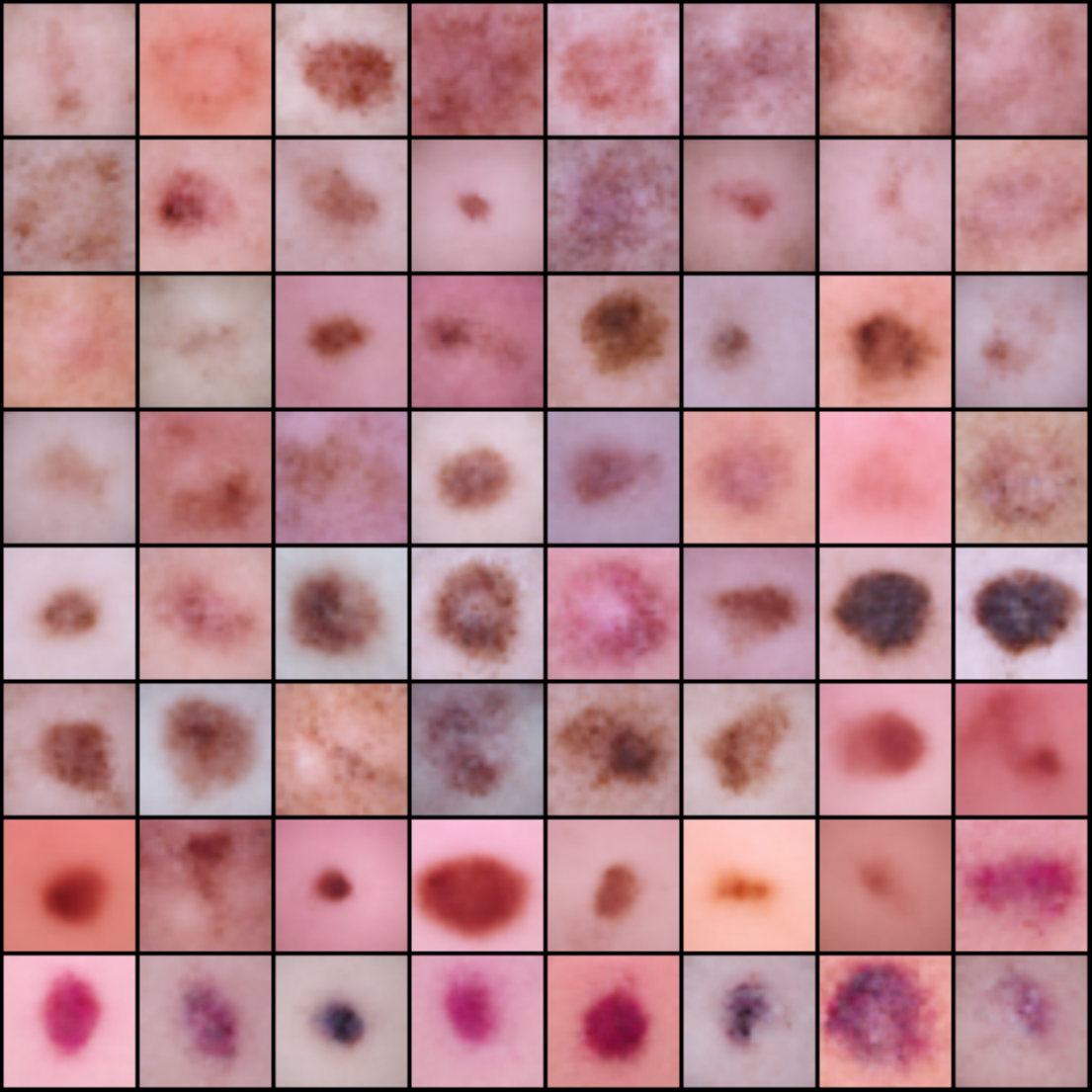} &
\includegraphics[width=0.16\textwidth]{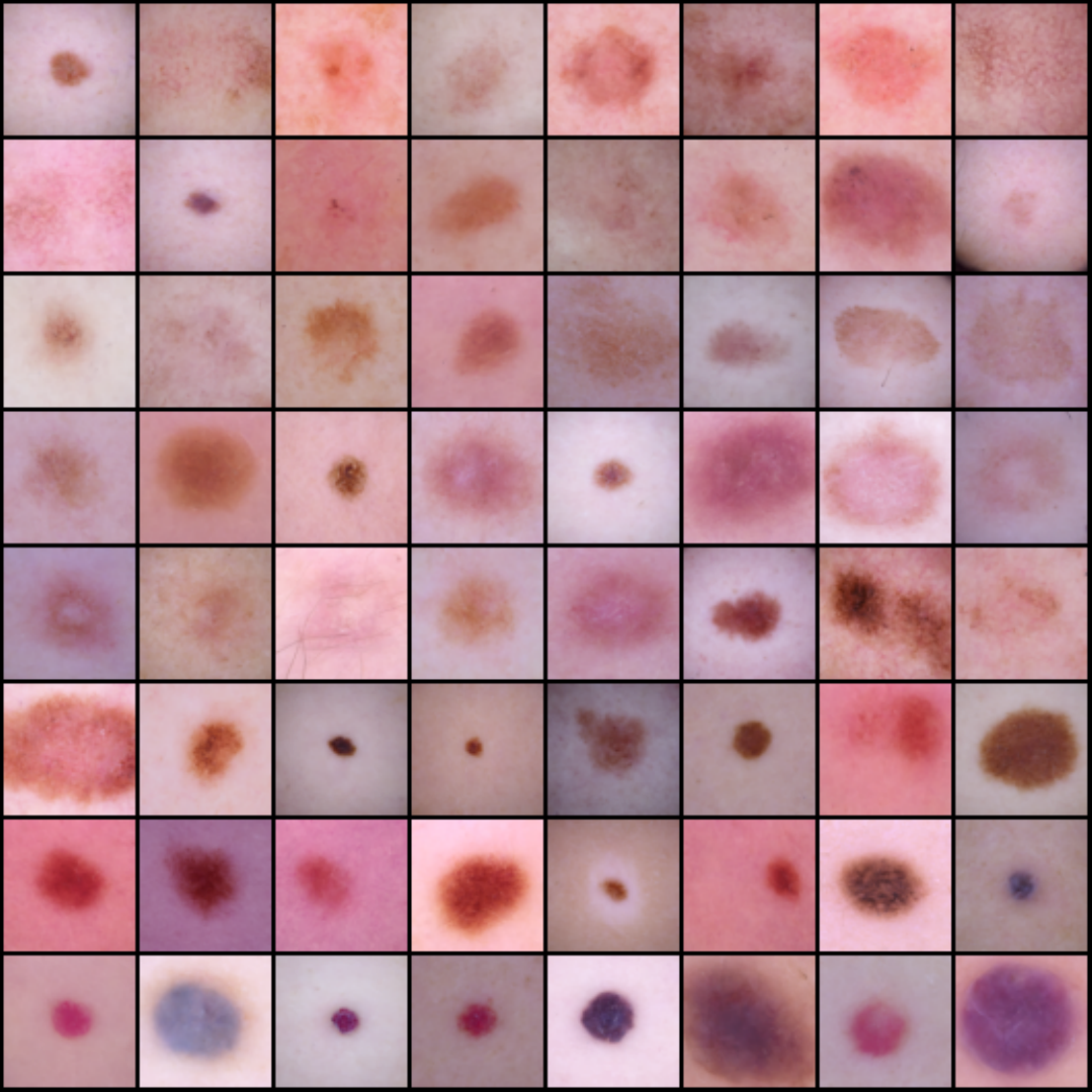} &
\includegraphics[width=0.16\textwidth]{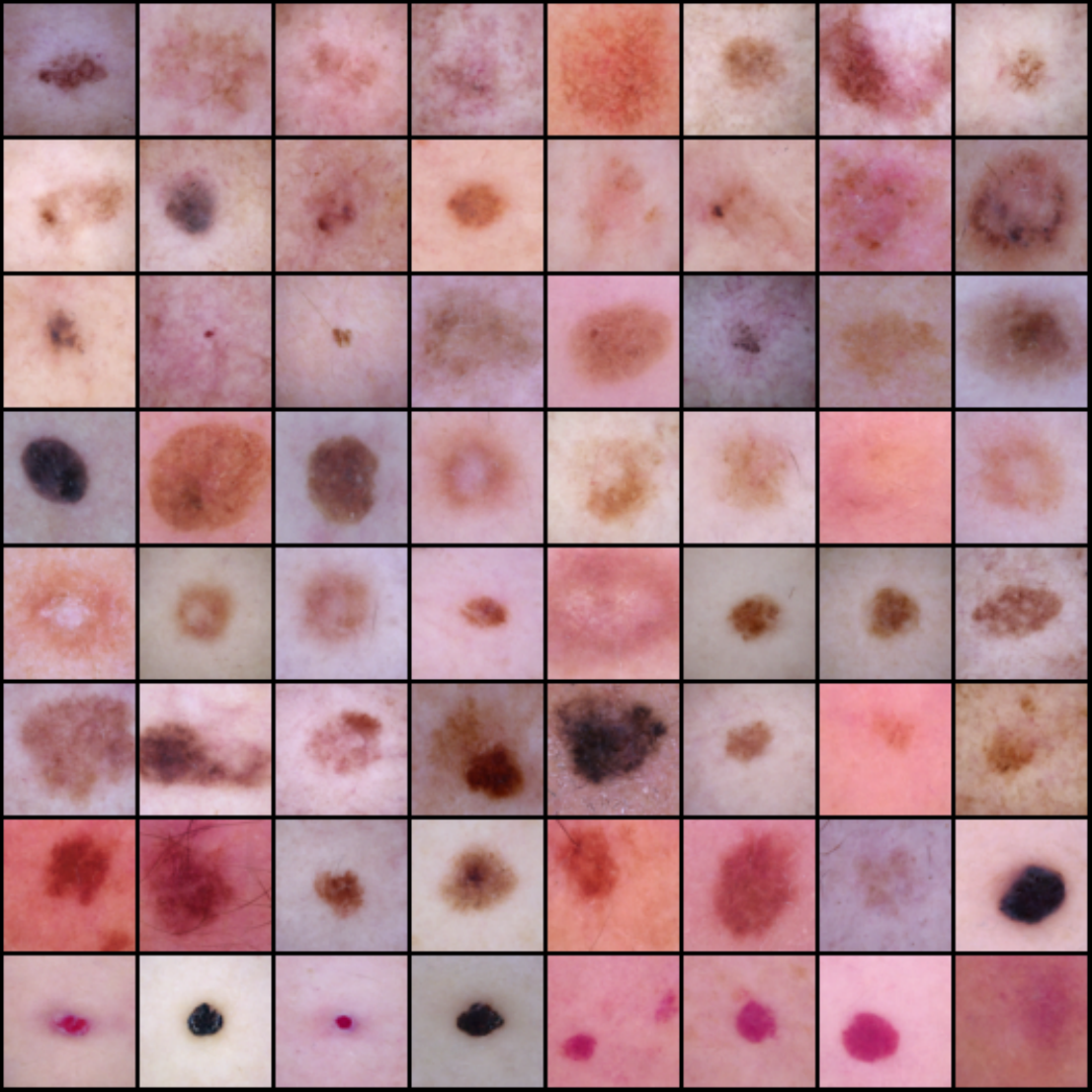} &
\includegraphics[width=0.16\textwidth]{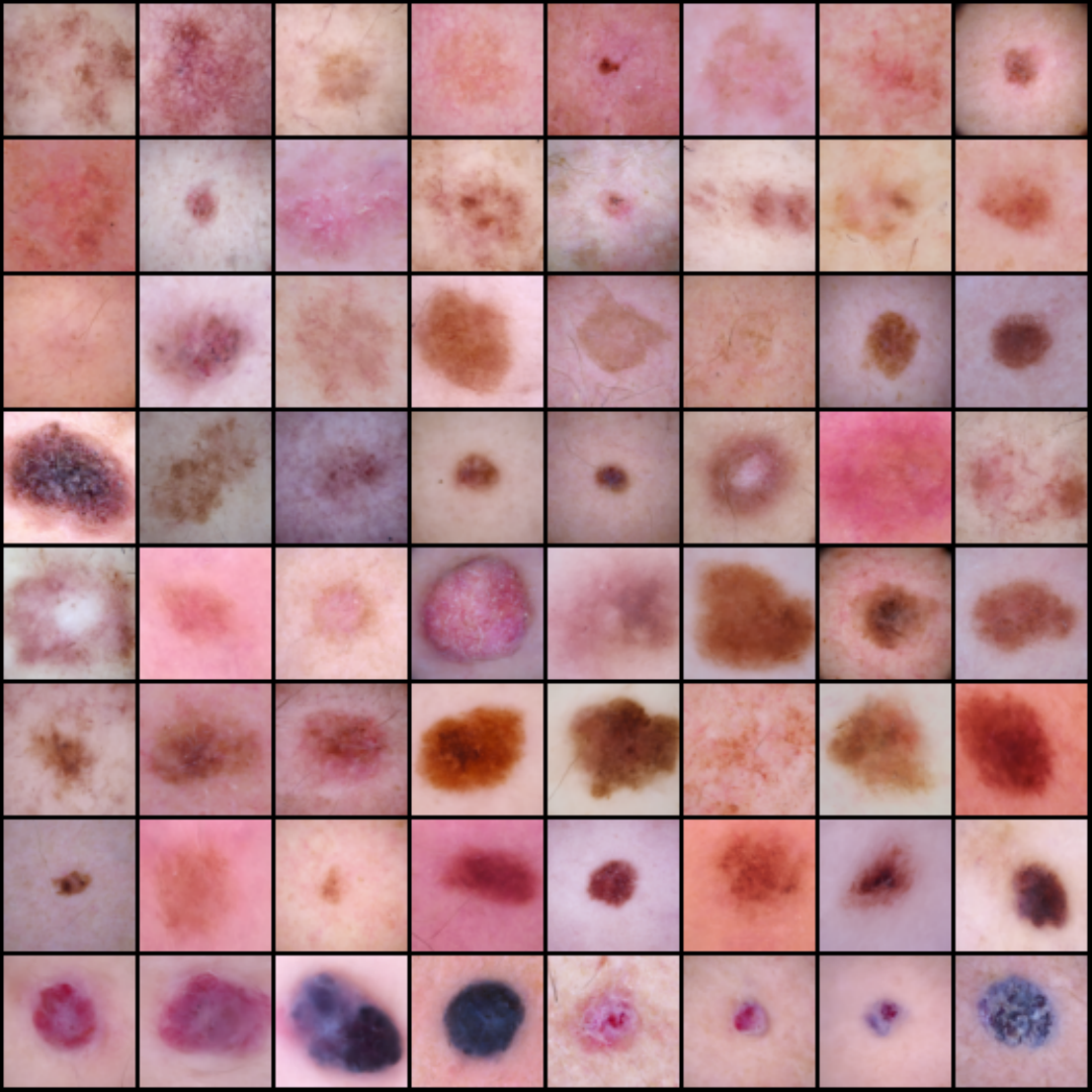} \\[2pt]
\bloodsamplerow
\parbox[b][0.16\textwidth][c]{1.1em}{\centering\rotatebox[origin=c]{90}{\small MNIST}} &
\includegraphics[width=0.16\textwidth]{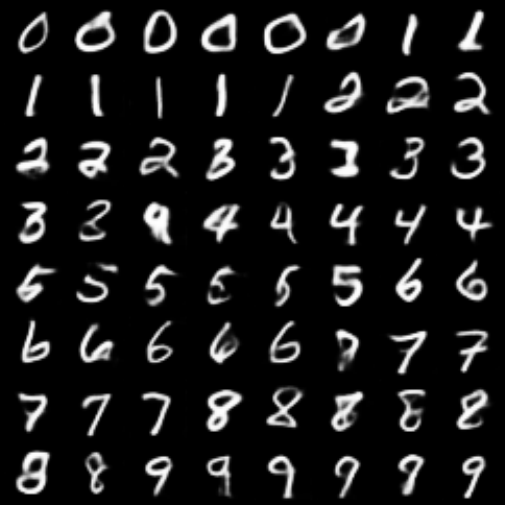} &
\includegraphics[width=0.16\textwidth]{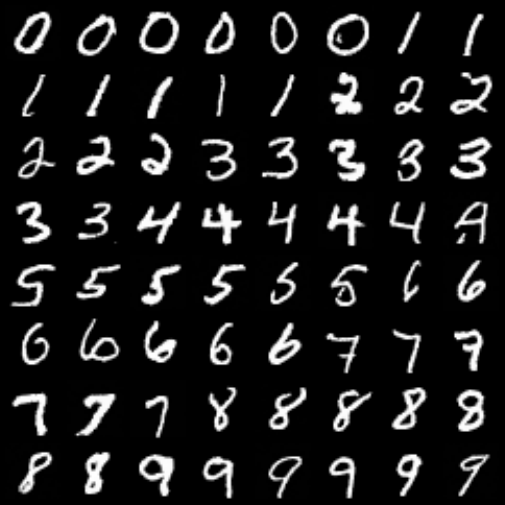} &
\includegraphics[width=0.16\textwidth]{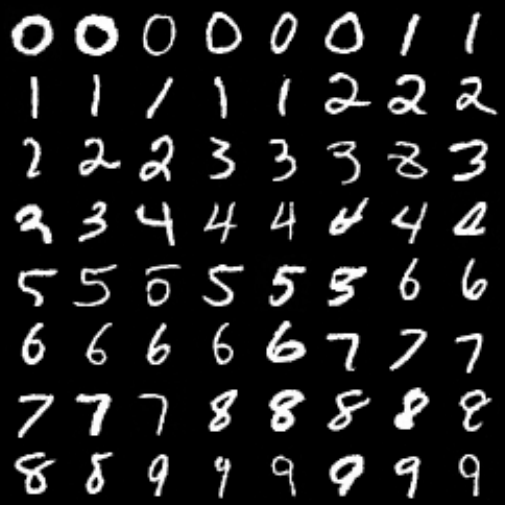} &
\includegraphics[width=0.16\textwidth]{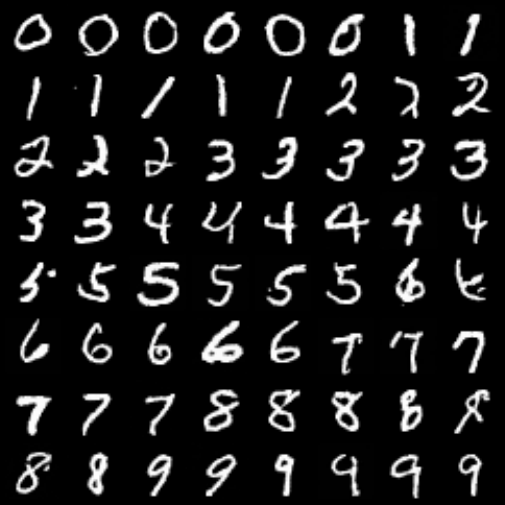} \\[2pt]
\parbox[b][0.16\textwidth][c]{1.1em}{\centering\rotatebox[origin=c]{90}{\small Fashion-MNIST}} &
\includegraphics[width=0.16\textwidth]{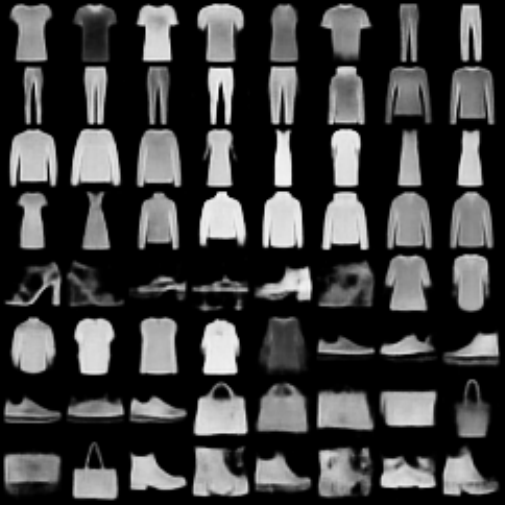} &
\includegraphics[width=0.16\textwidth]{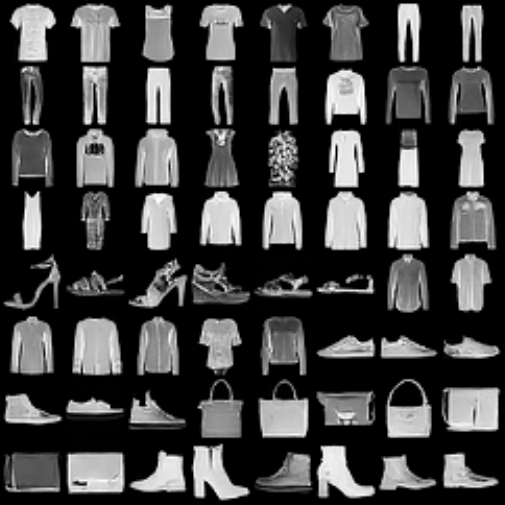} &
\includegraphics[width=0.16\textwidth]{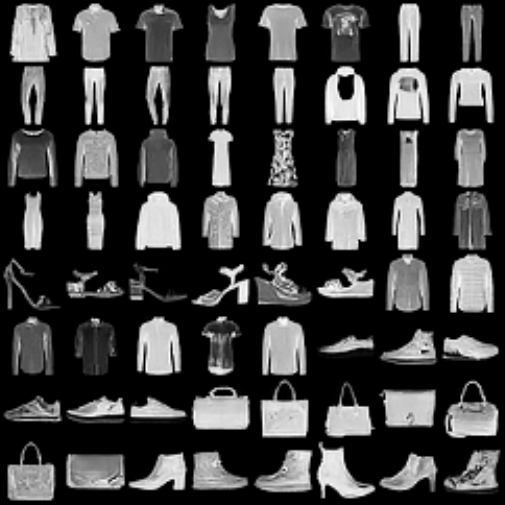} &
\includegraphics[width=0.16\textwidth]{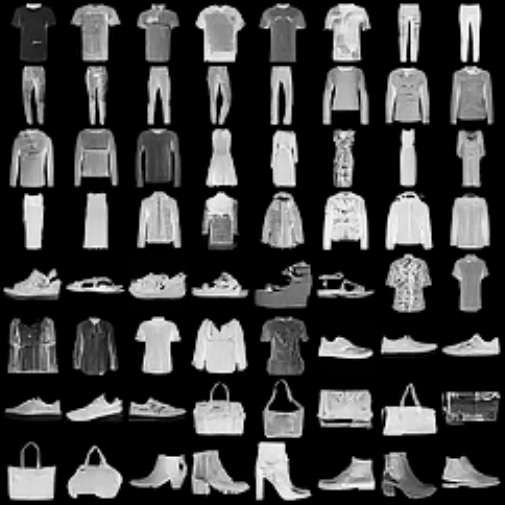} \\[2pt]
\parbox[b][0.16\textwidth][c]{1.1em}{\centering\rotatebox[origin=c]{90}{\small CIFAR-10}} &
\includegraphics[width=0.16\textwidth]{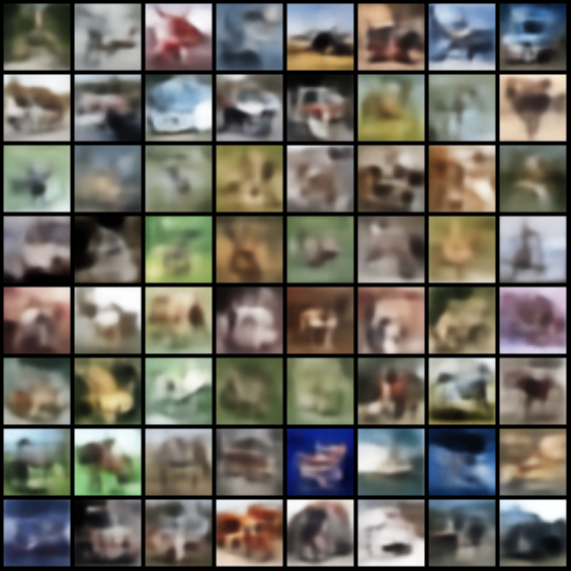} &
\includegraphics[width=0.16\textwidth]{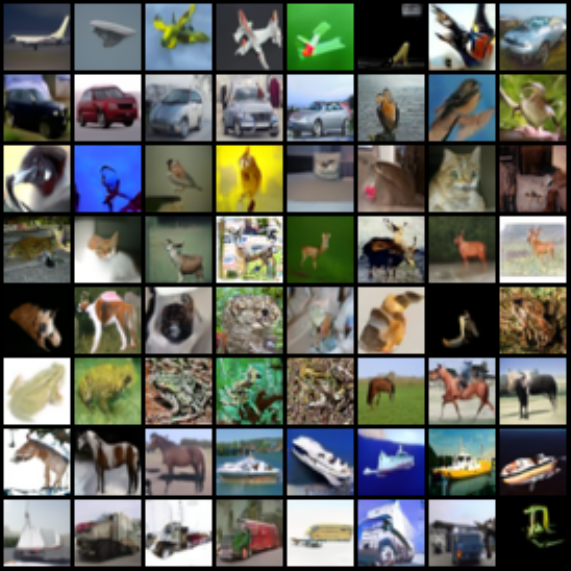} &
\includegraphics[width=0.16\textwidth]{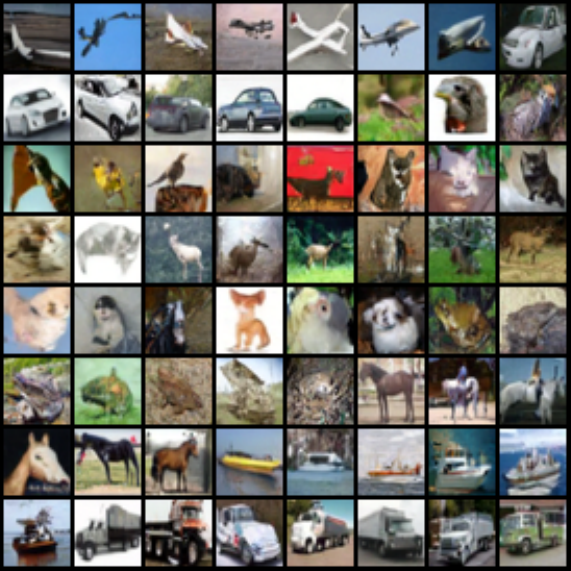} &
\includegraphics[width=0.16\textwidth]{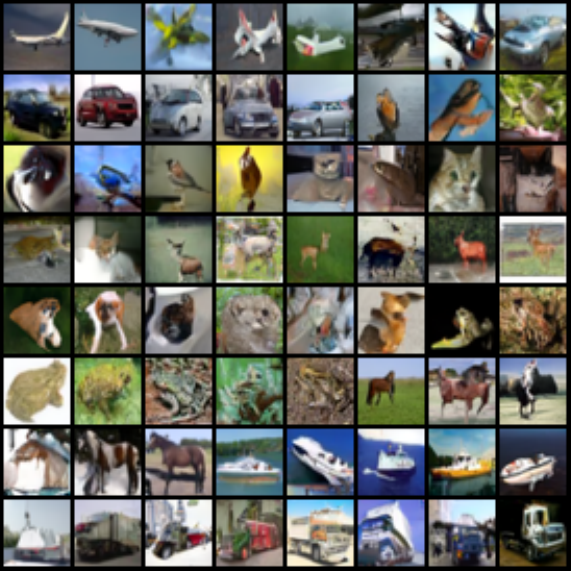} \\[2pt]
\parbox[b][0.16\textwidth][c]{1.1em}{\centering\rotatebox[origin=c]{90}{\small CIFAR-100}} &
\includegraphics[width=0.16\textwidth]{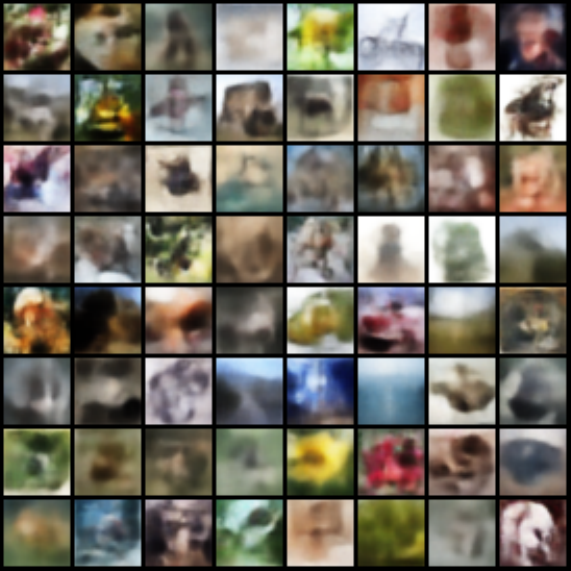} &
\includegraphics[width=0.16\textwidth]{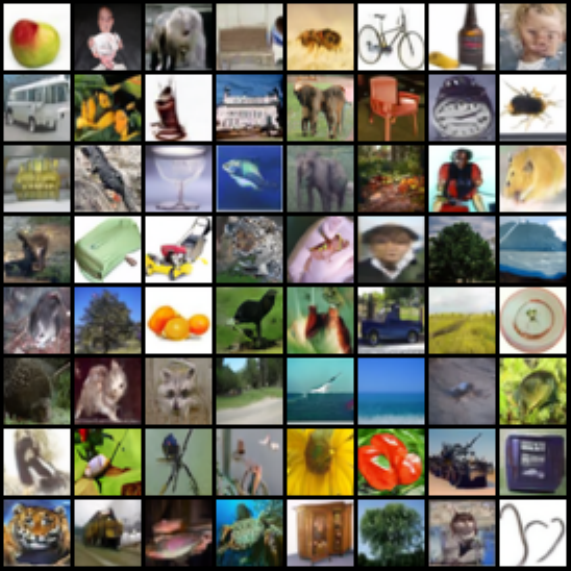} &
\includegraphics[width=0.16\textwidth]{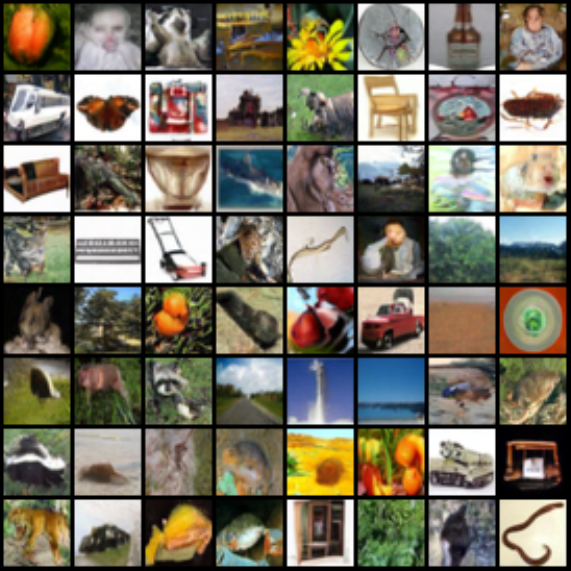} &
\includegraphics[width=0.16\textwidth]{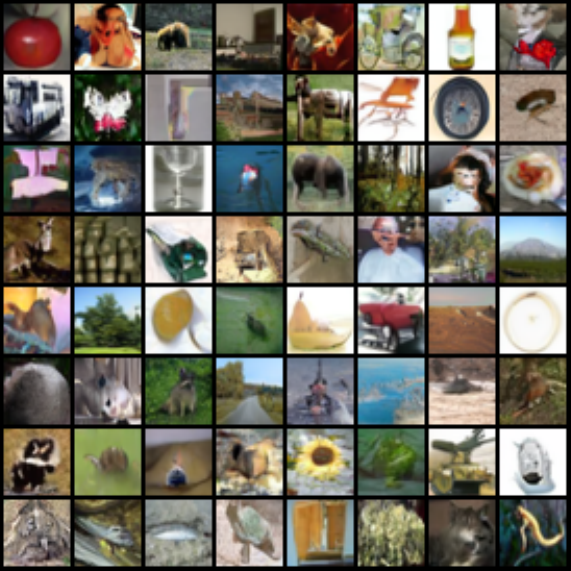} \\
\end{tabular}
\caption{{\marknew \textbf{Samples from every generator on \ifbloodsamples the six\else the five\fi\ datasets.} CIFAR-10: preview grids of the runs behind the synthetic datasets. Other rows: the same checkpoints with each generator's default sampler (DDPM: 1000 ancestral steps, not the 250 DDIM steps of its synthetic datasets).}}
\label{fig:samples-main}
\end{figure}

\clearpage
\section{Implementation and metric details}
\label{app:impl}

\begin{table}[h!]
\small
\setlength{\tabcolsep}{3pt}
\caption{\textbf{Experimental protocol.} $\alpha$: mixing ratio ($\alpha = 0$: real-only classifier, shared by all generators). $\rho$: real-data fraction.}
\label{tab:protocol}
\centering
\begin{tabular}{@{}ll@{}}
\toprule
Datasets & MNIST, Fashion-MNIST, CIFAR-10, \\
 & CIFAR-100, DermaMNIST, BloodMNIST \\
Classifiers & ResNet-18, ViT, Swin-T, ConvNeXt-V2 \\
Generators & VAE, DDPM, VP-SDE, flow matching; \\
 & traditional augmentation baseline (no generator) \\
Splits & official train/test, stratified tenth \\
 & of train held out for validation \\
Ratio $\alpha$ & $0.2, 0.4, 0.6, 0.8, 1.0$ at every $\rho$, \\
 & and $2.0, 4.0, 8.0$ at $\rho = 1$ \\
Fraction $\rho$ & $0.2, 0.4, 0.6, 0.8, 1.0$ \\
Augmentation & base: crop and flip (MNIST: none) \\
Sample quality & FID against the generator's own training set \\
\bottomrule
\end{tabular}
\end{table}

\paragraph{Design constraints in detail.}
Four constraints enforce the experimental design. First, each generator trains only on $\mathcal{D}_{\mathrm{train}}(\rho)$, so it sees nothing the classifier does not. We retrain it at every $\rho$, so no synthetic dataset leaks real images that the subsampling removed. Second, the validation set drives every model-selection choice,
{\marknew and each run uses the held-out test set once and never lets it inform a choice}. Third, within a dataset every generator receives the same number of image updates (epochs times training-set size), so a smaller generator training budget cannot explain a weaker downstream effect for one generator. Finally, we freeze the hyperparameters per backbone and dataset, so no configuration benefits from asymmetric tuning (\cref{app:hparams}).

\paragraph{What the design does and does not control.}
\leavevmode{\marknew Each backbone keeps its own fixed training recipe (\cref{tab:app-classifier})}, so configurations with different backbones also differ in training recipe, and ``the classifier'' means a backbone with its recipe. We fix the classifier's training budget in epochs, so raising $\alpha$ also raises the number of optimizer steps. \Cref{app:caveats} lists the couplings.

Synthetic images enter only the training mixture $\mathcal{D}_{\mathrm{mix}}(\rho, \alpha)$. We split the validation set $\mathcal{D}_{\mathrm{val}}$ off the official real training set, and the test set $\mathcal{D}_{\mathrm{test}}$ is the official real test set. No generated image enters either set, so a model fitting its own generator's artifacts can inflate neither checkpoint selection nor the reported accuracy.

\paragraph{Software and metrics.}
The code uses PyTorch \citep{paszke2019pytorch}, and each backbone comes from a standard package (\cref{tab:software}), with only the input changes of \cref{tab:app-classifier}. All four backbones start from random weights, and the only pretrained network is the InceptionV3 behind FID. Every classifier minimizes unweighted softmax cross-entropy, with neither label smoothing nor class weighting. The VAE, DDPM, the VP-SDE score model and their shared U-Net are our own code, and flow matching uses the \texttt{flow\_matching} library \citep{lipman2023flow}. FID comes from the \texttt{pytorch-fid} package \citep{heusel2017gans}. It compares a generator's full seed-0 synthetic dataset with that generator's own training set, capped at a stratified 10{,}000-image subsample, after mapping images to $[0, 1]$, repeating grayscale to three channels and resizing bilinearly to $299 \times 299$.

\begin{table*}[t]
\centering
\footnotesize
\caption{\textbf{Software, versions and sources.} Versions of every reported run. Sources: GitHub owner and repository. Only InceptionV3 in \texttt{pytorch-fid} loads pretrained weights.}
\label{tab:software}
\begin{tabular}{@{}lll@{}}
\toprule
Used for & Package and version & Source \\
\midrule
Tensors, training and sampling & PyTorch 2.8.0, Python 3.9, CUDA 12 & \href{https://github.com/pytorch/pytorch}{\texttt{pytorch/pytorch}} \\
ResNet-18, Swin-T, MNIST, Fashion-MNIST and CIFAR & torchvision 0.23.0 & \href{https://github.com/pytorch/vision}{\texttt{pytorch/vision}} \\
Crop, flip, affine and RandAugment & torchvision 0.23.0 & \href{https://github.com/pytorch/vision}{\texttt{pytorch/vision}} \\
ViT & \texttt{vit-pytorch} 1.24.1 & \href{https://github.com/lucidrains/vit-pytorch}{\texttt{lucidrains/vit-pytorch}} \\
ConvNeXt-V2 Femto & \texttt{timm} 1.0.28 & \href{https://github.com/huggingface/pytorch-image-models}{\texttt{huggingface/pytorch-image-models}} \\
Flow-matching path and ODE solver & \texttt{flow\_matching} 1.0.10 & \href{https://github.com/facebookresearch/flow_matching}{\texttt{facebookresearch/flow\_matching}} \\
VAE, DDPM, VP-SDE and U-Net & our code, in PyTorch & this paper \\
FID with InceptionV3 & \texttt{pytorch-fid} 0.3.0 & \href{https://github.com/mseitzer/pytorch-fid}{\texttt{mseitzer/pytorch-fid}} \\
DermaMNIST and BloodMNIST, 64 px & MedMNIST+ files & \citet{yang2023medmnist} \\
\bottomrule
\end{tabular}
\end{table*}

\paragraph{Preprocessing and batching.}
Real and synthetic images pass through the same pipeline. We map images to $[-1, 1]$, the range in which generators train and sample. Classifier loaders standardize each channel with the mean and standard deviation of the official training set. Training adds a pad-4 random crop and a horizontal flip, except on MNIST, and validation and test images receive only the standardization. The training mixture concatenates the real training set and the synthetic draw and is reshuffled at every epoch, with no fixed ratio within a batch.

\paragraph{Splits and seeds.}
The validation set is a stratified tenth of each class of the official training set, drawn once with split seed $0$ (\cref{tab:splits}). A real-data fraction $\rho$ keeps the first $\rho$ of each remaining class, so the subsets are nested and identical across run seeds. Generator training and each synthetic draw use seed $0$, so the three run seeds vary only the classifier's training and the class-balanced subsample of the synthetic dataset. Every configuration is a three-seed mean, and its standard deviation divides by the number of seeds minus one. Seeding fixes every data-side random choice, but we do not enable cuDNN determinism, so a rerun can differ in the last decimals.

\begin{table}[h!]
\centering
\small
\caption{\textbf{Image counts of the three splits.} Validation: a stratified tenth of the official training set. Training ($N$): the rest. Test: the official test set.}
\label{tab:splits}
\begin{tabular}{@{}lrrrr@{}}
\toprule
Dataset & Official training & Training ($N$) & Validation & Test \\
\midrule
MNIST & 60{,}000 & 54{,}000 & 6{,}000 & 10{,}000 \\
Fashion-MNIST & 60{,}000 & 54{,}000 & 6{,}000 & 10{,}000 \\
CIFAR-10 & 50{,}000 & 45{,}000 & 5{,}000 & 10{,}000 \\
CIFAR-100 & 50{,}000 & 45{,}000 & 5{,}000 & 10{,}000 \\
DermaMNIST & 7{,}007 & 6{,}306 & 701 & 2{,}005 \\
BloodMNIST & 11{,}959 & 10{,}763 & 1{,}196 & 3{,}421 \\
\bottomrule
\end{tabular}
\end{table}

\FloatBarrier

\section{Backbone and generator settings}
\label{app:float-settings}
\label{app:hparams}

{\marknew \Cref{tab:app-classifier,tab:app-backbone,tab:app-family} give the settings of the backbones, the shared generator network and each generator, and \cref{app:impl} gives the seed basis. Classifier training uses a cosine schedule, stepped once per epoch and decaying to zero without warmup or restarts, and no gradient clipping. Weight decay applies to every parameter, biases and normalization weights included. Checkpoint selection breaks validation-accuracy ties toward the earlier epoch. ResNet-18 and the ViT take the native resolution and channel count. Swin-T and ConvNeXt-V2 resize inside the model, so augmentation runs at the native resolution. Their fixed input adapter keeps the normalization, repeats grayscale to three channels and interpolates bilinearly to 64 px with corner alignment off, and DermaMNIST and BloodMNIST pass through at their native 64 px. The ViT uses dropout 0.1 in its transformer blocks and patch embedding. The other backbones keep their packages' defaults, which are stochastic depth rising linearly to 0.2 for Swin-T and no dropout or stochastic depth for ResNet-18 and ConvNeXt-V2.}

\begin{table}[h!]
\caption{\textbf{Backbone settings, identical for every generator and mixing ratio.}}
\label{tab:app-classifier}
\begin{center}
\scriptsize
\setlength{\tabcolsep}{3.1pt}
\begin{tabular}{lcccc}
\toprule
Setting & ResNet-18 & ViT & Swin-T & ConvNeXt-V2 \\
\midrule
input & native & native & 64 px & 64 px \\
stem / patch & 3 by 3 conv s1, no max pool & patch 4 (8 at 64 px) & patch 4, window 7 & 4 by 4 conv s4 \\
architecture & torchvision resnet18 & dim 256 / depth 6 / heads 8 / MLP 512 / dropout 0.1 & torchvision swin\_t & timm convnextv2\_femto \\
initialization & random & random & random & random \\
trained parameters & all & all & all & all \\
optimizer & SGD, momentum 0.9 & AdamW & AdamW & AdamW \\
learning-rate schedule & cosine & cosine & cosine & cosine \\
epochs & 30 (MNIST); 100 & 200 & 30 (MNIST); 100 & 100 \\
batch size & 128 & 128 & 128 & 128 \\
learning rate & 0.1 & 0.0003 & 0.0003 & 0.001 \\
weight decay & 0.0005 & 0.05 & 0.05 & 0.05 \\
base augmentation setting & \multicolumn{4}{c}{crop pad 4 + flip (CIFAR, Fashion-MNIST, DermaMNIST, BloodMNIST); none (MNIST); then normalize} \\
traditional augmentation baseline & \multicolumn{4}{c}{RandAugment \citep{cubuk2020randaugment}, 2 operations, magnitude 9 (MNIST, Fashion-MNIST: affine, 10 degrees / 0.1 translation)} \\
\bottomrule
\end{tabular}
\end{center}
\end{table}
\begin{table}[h!]
\caption{\textbf{Generator network and training settings per dataset.} Network rows: the shared U-Net. Training rows: all four generators.}
\label{tab:app-backbone}
\begin{center}
\scriptsize
\begin{tabular}{lcccc}
\toprule
Setting & MNIST & Fashion-MNIST & CIFAR-10/100 & DermaMNIST / BloodMNIST \\
\midrule
base channels & 32 & 32 & 128 & 128 \\
channel multipliers & (1, 2) & (1, 2) & (1, 2, 2, 2) & (1, 2, 2, 4) \\
residual blocks per resolution & 2 & 2 & 2 & 2 \\
attention resolutions & (16, 14) & (16, 14) & (16, 14) & (16, 14) \\
dropout & 0.1 & 0.1 & 0.1 & 0.1 \\
optimizer & AdamW & AdamW & AdamW & AdamW \\
learning-rate schedule & constant & constant & constant & constant \\
EMA decay & 0.999 & 0.999 & 0.999 & 0.999 \\
epochs & 30 & 167 & 200 & 1427 / 836 \\
image updates & 1.6M & 9.0M & 9.0M & 9.0M \\
batch size & 128 & 128 & 128 & 128 \\
learning rate & 0.0002 & 0.0002 & 0.0002 & 0.0002 \\
weight decay & 0.01 & 0.01 & 0.01 & 0.01 \\
gradient-norm clip & 1.0 & 1.0 & 1.0 & 1.0 \\
horizontal flip while training & no & yes & yes & yes \\
\bottomrule
\end{tabular}
\end{center}
\end{table}
\begin{table}[h!]
\caption{\textbf{Generator-specific settings, identical across datasets unless noted.}}
\label{tab:app-family}
\begin{center}
\scriptsize
\begin{tabular}{llc}
\toprule
Generator & Setting & Value \\
\midrule
DDPM & timesteps $T$ & 1000 \\
 & noise schedule & cosine \\
 & label-drop probability (CFG training) & 0.1 \\
 & guidance scale at sampling & 0.0 \\
 & sampler & DDIM, 250 steps \\
VAE & latent dimension & 128 ($28 \times 28$ datasets); 256 (others) \\
 & $\beta$ & 1.0 \\
VP-SDE & $\beta_{\min}$ / $\beta_{\max}$ & 0.1 / 20.0 \\
 & time cutoff $t_{\epsilon}$ & 0.001 \\
 & predictor steps & 500 \\
 & corrector steps per iteration & 1 \\
 & corrector signal-to-noise ratio & 0.16 \\
Flow matching & probability path & CondOT \\
 & ODE solver / steps & midpoint / 50 \\
\bottomrule
\end{tabular}
\end{center}
\end{table}

\FloatBarrier
\section{Generator losses and samplers}
\label{app:generators}

We build each generator of \cref{sec:instantiations} as follows. DDPM, the VP-SDE score model and flow matching share one U-Net. The U-Net conditions on a sinusoidal embedding of the time in $[0, 1]$, scaled by 1000 inside the embedding and mapped by a two-layer MLP to four times the base width. DDPM divides its integer step $t$ by its number of steps $T$, so all three generators feed the network a time in $[0, 1]$. Time $1$ is the noise end for DDPM and the VP-SDE and the data end for flow matching. The class enters as a learned embedding of the same width added to the time embedding. One extra null row of the embedding table stands in for labels dropped during classifier-free-guidance training (only DDPM drops labels, \cref{tab:app-family}). Residual blocks use GroupNorm with $\min(8, \text{number of channels})$ groups and SiLU and receive the summed embedding through a per-block linear projection. Self-attention is four-head scaled dot-product attention over spatial positions, also in the first bottleneck block. Downsampling is a stride-2 convolution and upsampling is nearest-neighbor interpolation followed by a $3 \times 3$ convolution. Each up stage runs one more residual block than its down stage, and the output convolution is zero-initialized.

\paragraph{Conditional VAE.}
The model learns an encoder $q_{\phi}(z \mid x, y)$ and a decoder $p_{\theta}(x \mid z, y)$ by minimizing the negative $\beta$-weighted evidence lower bound (ELBO) \citep{kingma2014auto, rezende2014stochastic}
\begin{equation}
\label{eq:vae}
\mathcal{L}_{\mathrm{VAE}}(\theta, \phi) = \E_{(x, y)} \bigl[ - \E_{q_{\phi}(z \mid x, y)} \log p_{\theta}(x \mid z, y) + \beta \, \KL\bigl( q_{\phi} \,\Vert\, \mathcal{N}(0, \mathbf{I}) \bigr) \bigr],
\end{equation}
with $\beta = 1.0$ and latent dimension 128 for the $28 \times 28$ datasets and 256 otherwise. Both terms are per-image sums (the squared reconstruction error over all pixels and the KL over latent dimensions) averaged over the batch, so $\beta = 1.0$ weighs them in the standard ELBO convention rather than against a per-pixel mean. The encoder is three stride-2 convolutions at one, two and four times the base channel count of \cref{tab:app-backbone}, each followed by GroupNorm and SiLU. The class enters the encoder as a learned per-class plane concatenated to the image as an extra channel and the decoder as a learned class vector concatenated with $z$. The decoder mirrors the encoder with transposed convolutions, resizes bilinearly to the native resolution where the mirrored size differs and ends in a tanh. Sampling draws $z \sim \mathcal{N}(0, \mathbf{I})$ and decodes with the requested label.

\paragraph{Denoising diffusion (DDPM).}
The model learns to predict the noise added by a fixed forward process (the $\epsilon$-parameterization) \citep{ho2020denoising}. With $\mu_t$ the signal scale of the noise schedule at step $t$ and $\sigma_t = \sqrt{1 - \mu_t^2}$ its noise scale, the loss is
\begin{equation}
\label{eq:ddpm}
\mathcal{L}_{\mathrm{DDPM}}(\theta) \;=\; \E_{(x, y),\, t,\, \epsilon} \Big[ \big\| \epsilon - \epsilon_{\theta}\big( \mu_t\, x + \sigma_t\, \epsilon,\ t,\ y \big) \big\|^2 \Big],
\end{equation}
with $\epsilon \sim \mathcal{N}(0, \mathbf{I})$ and $t$ uniform over the $T = 1000$ steps of a cosine schedule \citep{nichol2021improved}.
{\marknew In training, we average the squared errors of \cref{eq:ddpm,eq:sde,eq:cfm}, $\| \epsilon - \epsilon_{\theta}(\mu_t x + \sigma_t \epsilon, t, y) \|^2$ for DDPM and the VP-SDE model and $\| v_{\theta}((1 - t) \epsilon + t x, t, y) - (x - \epsilon) \|^2$ for flow matching, over pixels as well as over the batch, and the gradient-norm clip of \cref{tab:app-backbone} acts on that average.} Training drops the label with probability 0.1 (classifier-free guidance \citep{ho2022classifier}). We sample every DDPM synthetic dataset conditionally at guidance scale 0.0 with DDIM \citep{song2021denoising}, a deterministic accelerated sampler, in 250 steps. The sampler runs over an evenly spaced subset of the 1000 training steps with DDIM stochasticity $\eta = 0$, clamps the predicted clean image to $[-1, 1]$ at every step and returns that prediction at the final step.

\paragraph{Score-based model (VP-SDE).}
The model learns to reverse a variance-preserving stochastic differential equation, $\mathrm{d}x_t = -0.5\, \beta(t)\, x_t \, \mathrm{d}t + \sqrt{\beta(t)}\, \mathrm{d}w_t$ started from a data image $x$, with $w_t$ a standard Wiener process and $\beta(t)$ linear between $\beta_{\min} = 0.1$ and $\beta_{\max} = 20.0$ \citep{song2021score}. Its transition kernel is Gaussian, $x_t = \mu_t x + \sigma_t \epsilon$ with $\mu_t = \exp\bigl(-0.5 \int_{0}^{t} \beta(t')\, \mathrm{d}t'\bigr)$ and $\sigma_t = \sqrt{1 - \mu_t^2}$, so the model trains by denoising score matching in the $\epsilon$-parameterization
\begin{equation}
\label{eq:sde}
\mathcal{L}_{\mathrm{SDE}}(\theta) \;=\; \E_{(x, y),\, t,\, \epsilon} \Big[ \big\| \epsilon - \epsilon_{\theta}\big( \mu_t x + \sigma_t \epsilon,\ t,\ y \big) \big\|^2 \Big], \qquad s_{\theta}(x_t, t, y) \;=\; - \epsilon_{\theta}(x_t, t, y) / \sigma_t ,
\end{equation}
with $t$ uniform on $[0.001, 1]$ (the lower cutoff avoids the singular limit). Sampling integrates the reverse SDE with an Euler-Maruyama predictor and one Langevin corrector step per iteration (predictor-corrector sampling, 500 steps, signal-to-noise ratio 0.16). 
{\marknew The sampler times are 501 uniformly spaced points from $t = 1$ down to the cutoff. At each of the first 500 the corrector runs and then the predictor steps to the next, so the last step ends at the cutoff.} The corrector's Langevin step size is $2\, (1 - \beta(t)\, \Delta t)\, (0.16\, \lVert \epsilon \rVert / \lVert s_{\theta} \rVert)^2$, where $0.16$ is the signal-to-noise ratio, $\Delta t$ is the spacing of the sampler times and $\epsilon \sim \mathcal{N}(0, \mathbf{I})$ is the noise of the Langevin step. The noise and score norms are batch averages, and the final predictor step adds no noise.

\paragraph{Conditional flow matching.}
The model learns a velocity field that moves noise to data along straight conditional paths (the conditional optimal-transport path, CondOT) \citep{lipman2023flow}. With $\epsilon \sim \mathcal{N}(0, \mathbf{I})$ and a data image $x$, the path point $x_t = (1 - t)\, \epsilon + t\, x$ has target velocity $x - \epsilon$, and the loss is
\begin{equation}
\label{eq:cfm}
\mathcal{L}_{\mathrm{CFM}}(\theta) \;=\; \E_{(x, y),\, \epsilon,\, t} \Big[ \big\| v_{\theta}\big( (1 - t)\, \epsilon + t\, x,\ t,\ y \big) - (x - \epsilon) \big\|^2 \Big],
\end{equation}
with $t$ uniform on $[0, 1]$. Sampling integrates the learned ordinary differential equation with a 50-step midpoint solver.
\paragraph{Weight averaging.}
All four generators keep an exponential moving average (EMA) of their weights with decay 0.999 and sample with the EMA weights. We update the average after every optimizer step. The average starts from the initial weights and uses no warmup or bias correction.

\paragraph{The traditional augmentation baseline.}
This baseline replaces generated images with traditionally augmented real ones and uses no generative model. Each image is a real training image that we transform once, when we build the set. RandAugment (2 operations, magnitude 9) on the RGB datasets or the small affine of \cref{tab:app-classifier} on the $28 \times 28$ datasets comes first, then the pad-4 random crop and horizontal flip on every dataset but MNIST. Draws are class-balanced with replacement under seed 0, with equal per-class counts like every generated synthetic dataset $\mathcal{S}(\rho)$. During classifier training its images receive the same on-the-fly augmentation as any other training image.

\FloatBarrier
\section{FID on the full real training set}
\label{app:float-pools}
\begin{table}[h!]
\centering
\small
\caption{\textbf{FID of every synthetic dataset at $\rho = 1$.} Seed-0 draws against the full real training set (\cref{app:impl}). Trad.\ aug.: traditional augmentation. Bold: lowest per dataset.}
\label{tab:pools}
\begin{tabular}{@{}lrrrrrr@{}}
\toprule
Generator & MNIST & Fashion-MNIST & CIFAR-10 & CIFAR-100 & DermaMNIST & BloodMNIST \\
\midrule
VAE & 20.9 & 47.4 & 113.9 & 101.2 & 54.2 & 127.3 \\
DDPM & 5.0 & \textbf{4.3} & 11.0 & 9.4 & 33.3 & 5.3 \\
VP-SDE & 8.0 & 5.8 & \textbf{6.7} & \textbf{6.3} & 25.2 & 4.8 \\
Flow matching & \textbf{3.0} & 5.0 & 9.3 & 11.0 & \textbf{21.8} & \textbf{3.4} \\
Trad.\ aug. & 8.9 & 19.1 & 37.5 & 39.7 & 132.5 & 110.9 \\
\bottomrule
\end{tabular}
\end{table}

\FloatBarrier
\section{Notation}
\label{app:notation}
{\marknew \Cref{tab:notation} lists the symbols of \cref{sec:tv,app:tv}.}

\begingroup
\footnotesize
\renewcommand{\arraystretch}{0.92}
\begin{longtable}{@{}p{0.17\linewidth}p{0.65\linewidth}p{0.12\linewidth}@{}}
\caption{\textbf{Notation of \cref{sec:tv,app:tv}.} Hats mark estimates. Subscript $\alpha$: mixing ratio. Dependence on $\rho$ is implicit except in $G_{\rho}$.}
\label{tab:notation}\label{tab:tv-notation} \\
\toprule
Symbol & Meaning & Defined in \\
\midrule
\endfirsthead
\toprule
Symbol & Meaning & Defined in \\
\midrule
\endhead
\bottomrule
\endlastfoot
\multicolumn{3}{@{}l}{\emph{Data and distributions}} \\
$N$ & number of images in the full real training set $\mathcal{D}_{\mathrm{train}}(1)$ & \cref{sec:protocol} \\
$\rho$ & real-data fraction, the share of each class of $\mathcal{D}_{\mathrm{train}}(1)$ that $\mathcal{D}_{\mathrm{train}}(\rho)$ keeps & \cref{sec:protocol} \\
$n$ & number of real training points, $n = \rho N$ & \cref{sec:protocol} \\
$\alpha$ & mixing ratio, so $\alpha \rho N$ synthetic points join the $\rho N$ real ones & \cref{sec:protocol} \\
$P$ & true distribution of labeled images $(x, y)$ & \cref{sec:tv} \\
$(x_{i}, y_{i})$ & the $n$ points of $\mathcal{D}_{\mathrm{train}}(\rho)$, assumed independent draws from $P$ & \cref{sec:tv} \\
$G_{\rho}$, $G$ & distribution of the samples of a generator that we fit to $\mathcal{D}_{\mathrm{train}}(\rho)$, written $G$ when $\rho$ is fixed & \cref{sec:tv} \\
$P_{\alpha}$ & mixture $\frac{1}{1+\alpha} P + \frac{\alpha}{1+\alpha} G$, with synthetic share $\alpha/(1+\alpha)$ & \cref{sec:tv} \\
$K$ & number of classes & \cref{app:caveats} \\
$A$ & an event & \cref{sec:tv} \\
$d_{\mathrm{TV}}(P, G)$ & total variation between $P$ and $G$: $\sup_{A} | P(A) - G(A) |$ over events $A$ & \cref{sec:tv} \\
\midrule
\multicolumn{3}{@{}l}{\emph{Classifiers and risks}} \\
$\mathcal{F}$ & hypothesis class of the backbone & \cref{sec:tv} \\
$\ell$ & loss & \cref{sec:tv} \\
$M$ & bound on the loss: $\ell$ takes values in $[0, M]$; $M = 1$ for the 0-1 loss & \cref{sec:tv} \\
$g$ & loss of one classifier $f$: $g(x, y) = \ell(f(x), y)$ & \cref{app:tv-setting} \\
$R_{Q}(f)$ & risk of $f$ under a distribution $Q$: $\E_{(x, y) \sim Q}\, \ell(f(x), y)$ & \cref{sec:tv} \\
$\widehat{R}_{P_{\alpha}}(f)$ & average loss of $f$ on the $n + \alpha n$ training points & \cref{sec:tv} \\
$\widehat{R}^{\mathrm{test}}_{P}(f)$ & error of $f$ on the test set, an estimate of $R_{P}(f)$ & \cref{sec:tv} \\
$\widehat{R}^{\mathrm{test}}_{G_{\rho}}(f)$ & error of $f$ on about 5{,}000 fresh draws from $G_{\rho}$, an estimate of $R_{G_{\rho}}(f)$ & \cref{sec:results-read} \\
$\mathrm{Err}(\alpha)$, $\widehat{\mathrm{Err}}(\alpha)$ & population risk $R_{P}(\hat{f}_{\alpha})$ and its estimate, the test error $\widehat{R}^{\mathrm{test}}_{P}(\hat{f}_{\alpha})$ averaged over the three run seeds & \cref{sec:tv} \\
$\mathrm{Err}(\alpha) - \mathrm{Err}(0)$ & change in population risk from adding synthetic data, negative when the mixture gains, estimated by $\widehat{\mathrm{Err}}(\alpha) - \widehat{\mathrm{Err}}(0)$, minus the gain in test accuracy & \cref{sec:tv} \\
$L(G_{\rho})$ & largest $| R_{P}(f) - R_{G_{\rho}}(f) |$ over our classifiers trained at $\rho = 1$, a lower bound on $d_{\mathrm{TV}}(P, G_{\rho})$ & \cref{sec:results-read} \\
$\widehat{L}(G_{\rho})$ & estimated lower bound, the estimate of $L(G_{\rho})$ from $\widehat{R}^{\mathrm{test}}_{P}$ and $\widehat{R}^{\mathrm{test}}_{G_{\rho}}$ (\cref{tab:dtv}) & \cref{sec:results-read} \\
$\varepsilon$ & margin with $| \widehat{L}(G_{\rho}) - L(G_{\rho}) | \le \varepsilon$ with probability $0.95$ & \cref{tab:dtv} \\
$f^{\star}_{P}$ & best classifier in $\mathcal{F}$: minimizes $R_{P}$ over $\mathcal{F}$ & \cref{sec:tv} \\
$f^{\star}_{\alpha}$ & best classifier in $\mathcal{F}$ for the mixture: minimizes $R_{P_{\alpha}}$ over $\mathcal{F}$ & \cref{app:tv-setting} \\
$\mathcal{A}$ & training algorithm: maps a training set to a classifier in $\mathcal{F}$; it returns $\hat{f}_{\alpha}$ & \cref{sec:tv} \\
$\hat{f}_{\alpha}$ & classifier that $\mathcal{A}$ returns from the $n + \alpha n$ training points & \cref{sec:tv} \\
$\hat{f}_{0}$ & real-only classifier, $\hat{f}_{\alpha}$ at $\alpha = 0$ & \cref{sec:tv} \\
\midrule
\multicolumn{3}{@{}l}{\emph{Stability bound}} \\
$\delta$ & failure probability, in $(0, 1/2]$ & \cref{sec:tv} \\
$\beta_{m}$ & uniform-stability constant of $\mathcal{A}$ on $m$ points \citep[Definition~2.1]{zheng2023toward}; the bound uses $\beta = \beta_{(1+\alpha) \rho N}$ & \cref{sec:tv} \\
$\mathcal{T}$ & stability of the learned distribution \citep[Theorem~3.1]{zheng2023toward}: the largest change, in total variation, of the law of the $\alpha \rho N$ synthetic draws when we replace one point of $\mathcal{D}_{\mathrm{train}}(\rho)$, the supremum of $d_{\mathrm{TV}}(G_{\mathcal{D}}^{\alpha \rho N}, G_{\mathcal{D}^{(i)}}^{\alpha \rho N})$ over training sets $\mathcal{D}$, indices $i$ and replacement points & \cref{sec:tv} \\
$G_{\mathcal{D}}^{m}$ & law of $m$ independent draws from the generator fitted to the training set $\mathcal{D}$ & Supplementary \\
$\mathcal{D}^{(i)}$ & $\mathcal{D}$ with its $i$-th point replaced & Supplementary \\
$\lesssim$ & at most a universal constant times & Supplementary \\
\end{longtable}
\endgroup

\FloatBarrier
\section{Derivations and the test error}
\label{app:tv}

\subsection{Setting and the population-level bound}
\label{app:tv-setting}
Assume that the original training data $\mathcal{D}_{\mathrm{train}}=\{(x_i,y_i)\}_{i=1}^n$ are independent draws from the true distribution $P$. A deep generative model fitted to $\mathcal{D}_{\mathrm{train}}$ yields $G$, which serves as an approximation of $P$. In generative data augmentation, $\alpha n$ samples drawn independently from $G$ given $\mathcal{D}_{\mathrm{train}}$ (with fixed per-class counts, \cref{app:caveats}), where we assume that $\alpha n$ is an integer, serve as training data for classifiers $f\in\mathcal{F}$ together with $\mathcal{D}_{\mathrm{train}}$. Let
\begin{equation*}
    P_{\alpha}=\frac{n}{n+\alpha n}P+\frac{\alpha n}{n+\alpha n} G=\frac{1}{1+\alpha}P+\frac{\alpha}{1+\alpha}G.
\end{equation*}

\begin{lemma}[Total variation bound, \cref{eq:tv-gap}]
\label{lem:tv}
As in \citet{zheng2023toward}, assume that the loss function $\ell$ takes values in $[0,M]$. Then, for every classifier $f$, $|R_P(f)-R_{G}(f)|\leq M d_{\mathrm{TV}}(P,G)$, where $d_{\mathrm{TV}}(P,G)=\sup_A |P(A)-G(A)|$ is the total variation between $P$ and $G$, so that $\sup_{f\in\mathcal{F}}|R_P(f)-R_{G}(f)|\leq M d_{\mathrm{TV}}(P,G)$.
\end{lemma}
\begin{proof}
Let $z = (x, y)$ and $g(z)=\ell(f(x),y)$. As $g(z)\in[0,M]$ for all $z$, write $g(z)=\int_0^M \mathrm{1}_{\{g(z)>t\}}\mathrm{d}t$. Then, $R_P(f)=\E_P(g)=\int_0^M P\{g(z)>t\}\mathrm{d}t$ by Tonelli's theorem, likewise for $G$, and
\begin{equation*}
\begin{aligned}
    |R_P(f)-R_G(f)|&=|\E_P(g)-\E_G(g)|\\
    &=|\int_0^M [P\{g(z)>t\}-G\{g(z)>t\}]\mathrm{d}t|\\
    &\leq \int_0^M |P\{g(z)>t\}-G\{g(z)>t\}|\mathrm{d}t\\
    &\leq \int_0^M d_{\mathrm{TV}}(P,G) \mathrm{d}t \qquad \text{(each $\{g(z) > t\}$ is an event $A$)}\\
    &=M d_{\mathrm{TV}}(P,G) .
\end{aligned}
\end{equation*}
\end{proof}

\textbf{Everything at the population level.}

Let $R_P(f)=\E_{(x,y)\sim P}\{\ell(f(x),y)\}$ be the expected risk under the distribution $P$. Then, $f^{\star}_{P}=\arg\min_{f\in\mathcal{F}}R_P(f)$ and $f^{\star}_{\alpha}=\arg\min_{f\in\mathcal{F}}R_{P_{\alpha}}(f)$ are the best classifiers under the true distribution and the mixture, where we assume that both minima are attained. Since we test both classifiers under the true distribution $P$, we compare $R_{P}(f^{\star}_{P})$ and $R_{P}(f^{\star}_{\alpha})$, which are also at the population level.

\begin{proposition}[Population level]
\label{prop:case1}
If $\ell$ takes values in $[0, M]$, then
\begin{equation*}
    R_P(f^{\star}_{\alpha})-R_P(f^{\star}_{P})\leq \frac{2\alpha}{1+\alpha} M d_{\mathrm{TV}}(P,G).
\end{equation*}
\end{proposition}
\begin{proof}
\begin{equation*}
\begin{aligned}
    R_P(f^{\star}_{\alpha})-R_P(f^{\star}_{P})&=\{R_P(f^{\star}_{\alpha})-R_{P_{\alpha}}(f^{\star}_{\alpha})\}+\{R_{P_{\alpha}}(f^{\star}_{\alpha})-R_{P_{\alpha}}(f^{\star}_{P})\}+\{R_{P_{\alpha}}(f^{\star}_{P})-R_P(f^{\star}_{P})\}\\
    &\leq 2\sup_{f\in\mathcal{F}}|R_P(f)-R_{P_{\alpha}}(f)|\\
    &=2\sup_{f\in\mathcal{F}}|R_P(f)-\frac{1}{1+\alpha}R_{P}(f)-\frac{\alpha}{1+\alpha}R_{G}(f)|\\
    &=\frac{2\alpha}{1+\alpha}\sup_{f\in\mathcal{F}}|R_P(f)-R_{G}(f)| \\
    &{}\le \frac{2\alpha}{1+\alpha} M d_{\mathrm{TV}}(P,G) .
\end{aligned}
\end{equation*}
The first inequality holds because the middle bracket is at most $0$, since $f^{\star}_{\alpha}$ minimizes $R_{P_{\alpha}}$ over $\mathcal{F}$ and $f^{\star}_{P} \in \mathcal{F}$, and each outer bracket is at most the supremum. The next line uses $R_{P_{\alpha}}(f) = \frac{1}{1+\alpha} R_{P}(f) + \frac{\alpha}{1+\alpha} R_{G}(f)$, and the last uses \cref{lem:tv}, $|R_P(f)-R_{G}(f)|\leq M d_{\mathrm{TV}}(P,G)$ for every $f$.
\end{proof}

\subsection{Proof of the bound}
\label{app:tv-dep-case}

\Cref{sec:tv} states the bound, and this subsection proves it. We write $n = \rho N$ for the number of real points and $G$ for $G_{\rho}$, so the $\alpha n$ generated points and the $n + \alpha n$ training points are those of \cref{sec:tv}. This is the setting of \citet[Section~2.2]{zheng2023toward} in our notation, with the $n$ real points of $\mathcal{D}_{\mathrm{train}}$ drawn independently from $P$, the $\alpha n$ points drawn independently from $G$ given $\mathcal{D}_{\mathrm{train}}$, their mixture $P_{\alpha}$ and the classifier $\hat{f}_{\alpha}$ trained on all $n + \alpha n$ points, with our fixed per-class counts (\cref{app:caveats}). We condition on the generator's training randomness, which they ignore, and on the classifier's training seed and the validation set, so that $\mathcal{A}$ is a deterministic map from training sets to classifiers, as their Definition~2.1 requires. We also take $n \ge 2$, so that $\log n > 0$.

Since the best classifier for the mixture, $f^{\star}_{\alpha} = \arg\min_{f \in \mathcal{F}} R_{P_{\alpha}}(f)$, depends on $\mathcal{D}_{\mathrm{train}}$ through $G$, we compare $\hat{f}_{\alpha}$ with the fixed $f^{\star}_{P}$ and split
\begin{equation}
\label{eq:excess-split}
\begin{aligned}
R_{P}(\hat{f}_{\alpha}) - R_{P}(f^{\star}_{P}) ={}& \bigl\{ R_{P}(\hat{f}_{\alpha}) - \widehat{R}_{P_{\alpha}}(\hat{f}_{\alpha}) \bigr\} + \bigl\{ \widehat{R}_{P_{\alpha}}(\hat{f}_{\alpha}) - \widehat{R}_{P_{\alpha}}(f^{\star}_{P}) \bigr\} \\
&+ \bigl\{ \widehat{R}_{P_{\alpha}}(f^{\star}_{P}) - R_{P}(f^{\star}_{P}) \bigr\} .
\end{aligned}
\end{equation}

\begin{theorem}[\cref{eq:tv-stab-dep}]
\label{thm:dep}
In the setting of \cref{sec:tv}, assume that the training algorithm $\mathcal{A}$ is $\beta_{n + \alpha n}$-uniformly stable \citep[Definition~2.1]{zheng2023toward} and that $\ell$ takes values in $[0, M]$. Then, for any $\alpha \ge 0$ and any $\delta \in (0, 1/2]$, with probability at least $1 - \delta$,
\begin{equation*}
\begin{aligned}
R_{P}(\hat{f}_{\alpha}) - R_{P}(f^{\star}_{P}) \lesssim{}& \frac{2\alpha}{1+\alpha}\, M\, d_{\mathrm{TV}}(P, G) \\
&+ \Bigl( \frac{M}{\sqrt{n + \alpha n}} + \frac{\sqrt{\alpha n}}{1+\alpha}\, \beta_{n + \alpha n} \Bigr) \sqrt{\log\tfrac{1}{\delta}} + \frac{\beta_{n + \alpha n} \bigl( \log n + \alpha \log(\alpha n) \bigr) + M\, \mathcal{T} \log n}{1+\alpha}\, \log\tfrac{1}{\delta} \\
&+ \widehat{R}_{P_{\alpha}}(\hat{f}_{\alpha}) ,
\end{aligned}
\end{equation*}
where we take $\alpha \log(\alpha n) = 0$ at $\alpha = 0$.
\end{theorem}
\begin{proof}
\Citet[Theorem~3.1]{zheng2023toward} bound $| R_{P}(\hat{f}) - \widehat{R}_{P_{\alpha}}(\hat{f}) |$ for the classifier $\hat{f}$ that a uniformly stable algorithm returns from $n$ real points and $\alpha n$ points drawn from a generator fitted to them. We apply their theorem twice, once to $\mathcal{A}$ and once to an algorithm that ignores its data, and bound the three brackets of \cref{eq:excess-split} in turn.

The first bracket is the gap of their theorem for $\mathcal{A}$. Applied at $\delta/2$, the theorem gives, with probability at least $1 - \delta/2$,
\begin{equation*}
\begin{aligned}
R_{P}(\hat{f}_{\alpha}) - \widehat{R}_{P_{\alpha}}(\hat{f}_{\alpha}) \lesssim{}& \frac{\alpha}{1+\alpha} M\, d_{\mathrm{TV}}(P, G) + \Bigl( \frac{M}{\sqrt{n + \alpha n}} + \frac{\sqrt{\alpha n}}{1+\alpha}\, \beta_{n + \alpha n} \Bigr) \sqrt{\log\tfrac{1}{\delta}} \\
&+ \frac{\beta_{n + \alpha n} \bigl( \log n + \alpha \log(\alpha n) \bigr) + M\, \mathcal{T} \log n}{1+\alpha}\, \log\tfrac{1}{\delta} .
\end{aligned}
\end{equation*}
The display writes $M / \sqrt{n + \alpha n}$ for their $M (\sqrt{n} + \sqrt{\alpha n}) / (n + \alpha n)$, which is at most $\sqrt{2}$ times larger since $(1 + \sqrt{\alpha})^{2} \le 2 (1+\alpha)$, and $\log(1/\delta)$ for their $\log(2/\delta)$, which is at most twice as large since $\delta \le 1/2$.

The middle bracket is at most $\widehat{R}_{P_{\alpha}}(\hat{f}_{\alpha})$, the training error, since the loss is nonnegative and so $\widehat{R}_{P_{\alpha}}(f^{\star}_{P}) \ge 0$. It is at most $0$ when $\hat{f}_{\alpha}$ minimizes $\widehat{R}_{P_{\alpha}}$ over $\mathcal{F}$.

The third bracket is the gap of their theorem for the algorithm that returns $f^{\star}_{P}$ for every training set. This algorithm is $0$-uniformly stable, since its output never changes. The theorem with $\beta = 0$, applied at $\delta/2$, gives, with probability at least $1 - \delta/2$,
\begin{equation*}
\widehat{R}_{P_{\alpha}}(f^{\star}_{P}) - R_{P}(f^{\star}_{P}) \lesssim \frac{\alpha}{1+\alpha} M\, d_{\mathrm{TV}}(P, G) + \frac{M}{\sqrt{n + \alpha n}} \sqrt{\log\tfrac{1}{\delta}} + \frac{M\, \mathcal{T} \log n}{1+\alpha}\, \log\tfrac{1}{\delta} .
\end{equation*}

By a union bound, both bounds hold with probability at least $1 - \delta$. On this event, \cref{eq:excess-split} and the bounds on its three brackets give
\begin{equation*}
\begin{aligned}
R_{P}(\hat{f}_{\alpha}) - R_{P}(f^{\star}_{P}) \lesssim{}& \frac{\alpha}{1+\alpha} M\, d_{\mathrm{TV}}(P, G) + \Bigl( \frac{M}{\sqrt{n + \alpha n}} + \frac{\sqrt{\alpha n}}{1+\alpha}\, \beta_{n + \alpha n} \Bigr) \sqrt{\log\tfrac{1}{\delta}} \\
&+ \frac{\beta_{n + \alpha n} \bigl( \log n + \alpha \log(\alpha n) \bigr) + M\, \mathcal{T} \log n}{1+\alpha}\, \log\tfrac{1}{\delta} \\
&+ \widehat{R}_{P_{\alpha}}(\hat{f}_{\alpha}) \\
&+ \frac{\alpha}{1+\alpha} M\, d_{\mathrm{TV}}(P, G) + \frac{M}{\sqrt{n + \alpha n}} \sqrt{\log\tfrac{1}{\delta}} + \frac{M\, \mathcal{T} \log n}{1+\alpha}\, \log\tfrac{1}{\delta} \\
\lesssim{}& \frac{2\alpha}{1+\alpha}\, M\, d_{\mathrm{TV}}(P, G) + \Bigl( \frac{M}{\sqrt{n + \alpha n}} + \frac{\sqrt{\alpha n}}{1+\alpha}\, \beta_{n + \alpha n} \Bigr) \sqrt{\log\tfrac{1}{\delta}} \\
&+ \frac{\beta_{n + \alpha n} \bigl( \log n + \alpha \log(\alpha n) \bigr) + M\, \mathcal{T} \log n}{1+\alpha}\, \log\tfrac{1}{\delta} + \widehat{R}_{P_{\alpha}}(\hat{f}_{\alpha}) .
\end{aligned}
\end{equation*}
After the first $\lesssim$, the first two lines bound the first bracket, the third line bounds the middle bracket and the fourth line bounds the third bracket. The second step adds the two terms in $d_{\mathrm{TV}}(P, G)$ into $\frac{2\alpha}{1+\alpha} M\, d_{\mathrm{TV}}(P, G)$, and the second copies of the terms in $M / \sqrt{n + \alpha n}$ and in $\mathcal{T}$ at most double the first ones, which $\lesssim$ absorbs. This is the theorem. At $\alpha = 0$ there are no generated points, so $\mathcal{T} = 0$ and the terms in $d_{\mathrm{TV}}(P, G)$, $\sqrt{\alpha n}$ and $\alpha \log(\alpha n)$ vanish, and their Theorem~3.1 reduces to their Theorem~2.1 \citep[Section~3.1]{zheng2023toward}, so the same proof holds. With $n = \rho N$ the theorem is \cref{eq:tv-stab-dep}.
\end{proof}

\subsection{The test error}
\label{app:tv-sim}

{\marknew Each classifier is the best-validation checkpoint of a run with seed $0$, $1$ or $2$. With the 0-1 loss, $M = 1$ and each risk is an error rate. We estimate the population risk $R_{P}(\hat{f}_{\alpha})$ by $\widehat{R}^{\mathrm{test}}_{P}(\hat{f}_{\alpha})$, the error of $\hat{f}_{\alpha}$ on the test set. The official test set is a sample from $P$ that no generator or classifier sees during training or model selection, so this error is an unbiased estimate of $R_{P}(\hat{f}_{\alpha})$. \Cref{fig:interaction,fig:rho-blocks,fig:cost} average it over the three run seeds, which gives $\widehat{\mathrm{Err}}(\alpha)$ of \cref{sec:tv}. Its standard error is at most $0.005$ on the four benchmarks, whose test sets hold $10{,}000$ images each, $0.009$ on BloodMNIST and $0.011$ on DermaMNIST (\cref{tab:splits}). }

\end{document}